\documentclass[journal]{IEEEtran}

\usepackage[utf8]{inputenc} 
\usepackage[T1]{fontenc}    
\usepackage{hyperref}       
\usepackage{url}            
\usepackage{booktabs}       
\usepackage{amsfonts}       
\usepackage{nicefrac}       
\usepackage{microtype}      
\usepackage{xcolor}         

\usepackage{wrapfig}
\usepackage[utf8]{inputenc} 
\usepackage[T1]{fontenc}    
\usepackage{hyperref}       
\usepackage{url}            
\usepackage{booktabs}       
\usepackage{amsfonts}       
\usepackage{nicefrac}       
\usepackage{microtype}      
\usepackage{xcolor}         
\usepackage{latexsym}
\usepackage{amssymb}
\usepackage{amsmath}
\usepackage{booktabs}
\usepackage{enumerate}
\usepackage{graphicx}
\usepackage{subfigure}
\usepackage{xspace}
\usepackage{float}
\usepackage{bbm}
\usepackage{bm}
\usepackage{multirow}
\usepackage{booktabs}
\usepackage{color}
\usepackage{framed}
\usepackage{stfloats}
\usepackage{iitem}
\usepackage{amssymb}
\usepackage{pifont}
\usepackage{enumitem}
\usepackage{colortbl}
\usepackage{arydshln}
\usepackage{multirow}
\usepackage{multicol}

\usepackage[T1]{fontenc}
\usepackage{colortbl}
\usepackage{amsmath}
\usepackage{algorithm}
\usepackage{algpseudocode}
\definecolor{maroon}{cmyk}{0,0.87,0.68,0.32}

\newcommand{\gray}{\rowcolor[gray]{.90}}

\AtBeginDocument{%
  \providecommand\BibTeX{{%
    \normalfont B\kern-0.5em{\scshape i\kern-0.25em b}\kern-0.8em\TeX}}}
\hypersetup{colorlinks,linkcolor={blue},citecolor={blue},urlcolor={purple}}  
\ifCLASSOPTIONcompsoc
  \usepackage[nocompress]{cite}
\else
  \usepackage{cite}
\fi

\ifCLASSINFOpdf
\else
\fi

\begin{document}
%
\title{Exploring and Enhancing the Transfer of Distribution in Knowledge Distillation for Autoregressive Language Models}
%
%
%
%

%
\author{Jun~Rao,
        Xuebo Liu$^\dag$,
        Zepeng Lin,
        Liang~Ding,
        Jing Li, 
Dacheng~Tao,~\IEEEmembership{Fellow,~IEEE}
~and Min Zhang
\thanks{J. Rao, X. Liu, Z. Lin, J. Li and M. Zhang are with the Harbin Institute of Technology, 518000,  Shenzhen, China; L. Ding is with the School of Computer Science, University of Sydney, NSW 2006, Australia; D. Tao is with the College of Computing \& Data Science at Nanyang
Technological University. (email: \{rao7jun, zepenglin11, liangding.liam\}@gmail.com, \{liuxuebo, li.jing, zhangmin2021\}@hit.edu.cn, dacheng.tao@ntu.edu.sg)}
\thanks{$\dag$~Corresponding Author.}
}

\markboth{Journal of \LaTeX\ Class Files,~Vol.~14, No.~8, August~2015}%
{Shell \MakeLowercase{\textit{et al.}}: Bare Advanced Demo of IEEEtran.cls for IEEE Computer Society Journals}
\maketitle
\begin{abstract}
Knowledge distillation (KD) is a technique that compresses large teacher models by training smaller student models to mimic them.
The success of KD in auto-regressive language models mainly relies on Reverse KL for mode-seeking and student-generated output (SGO) to combat exposure bias.
Our theoretical analyses and experimental validation reveal that while Reverse KL effectively mimics certain features of the teacher distribution, it fails to capture most of its behaviors.
Conversely, SGO incurs higher computational costs and presents challenges in optimization, particularly when the student model is significantly smaller than the teacher model. 
These constraints are primarily due to the immutable distribution of the teacher model, which fails to adjust adaptively to models of varying sizes. 
We introduce Online Knowledge Distillation (OKD), where the teacher network integrates small online modules to concurrently train with the student model. 
This strategy abolishes the necessity for on-policy sampling and merely requires minimal updates to the parameters of the teacher's online module during training, thereby allowing dynamic adaptation to the student's distribution to make distillation better. 
Extensive results across multiple generation datasets show that OKD achieves or exceeds the performance of leading methods in various model architectures and sizes, reducing training time by up to fourfold. 
\end{abstract}

\begin{IEEEkeywords}
Knowledge Distillation, Large Language Model, Autoregressive Language Model.
\end{IEEEkeywords}

\IEEEdisplaynontitleabstractindextext

%
\IEEEpeerreviewmaketitle

\section{Introduction} 


\IEEEPARstart{W}{ith} the advancement of large language models (LLMs)~\cite{gpt3,gpt4,llama}, one major approach to reduce the computational cost of model inference is knowledge distillation (KD)~\cite{DBLP:journals/corr/HintonVD15}.
This process involves supervising a relatively smaller-scale student model with a larger teacher model to achieve a more efficient yet effective small model~\cite{10476767,10145833}. 
Prior to the era of LLMs, most knowledge distillation efforts were focused on non-autoregressive models such as BERT~\cite{PKD,DBLP:conf/cikm/RaoQQW0021,zhong2024panda} and CNNs~\cite{ProKD,KDCL} for classification tasks. 
However, in the current landscape, the majority of models, such as LLaMA~\cite{llama}, are based on the autoregressive transformer~\cite{DBLP:journals/corr/BahdanauCB14/self-attention,DBLP:conf/nips/VaswaniSPUJGKP17} architecture for generation tasks.
 
In autoregressive language models, exposure bias~\cite{bengio2015scheduled,arora-etal-2022-exposure} describes the overfitting issue that arises from divergent data distributions between training and testing phases. 
This bias inhibits the student model's ability to generate tokens across varied distributions accurately. 
Recent investigations~\cite{agarwal2023gkd,gu2023minillm} indicate that reverse Kullback-Leibler divergence (RKL) better directs models to learn specific patterns compared to forward KL (FKL). 
Moreover, modern on-policy KD strategies~\cite{gu2023minillm,agarwal2023gkd,wen-etal-2023-f} have shown efficacy in mitigating exposure bias through the use of student-generated tokens. 
Nonetheless, methods like RKL predominantly optimize difficult samples and student-generated output underperforms when significant disparities exist between the sizes of teacher and student networks, also prolonging training times. 
Despite these efforts, KD methods~\cite{DBLP:journals/corr/HintonVD15,agarwal2023gkd,wen-etal-2023-f} in auto-regressive LMs still face a mismatch in distributions between teacher and student models~\cite{wen-etal-2023-f}, contributing to an under-specification problem~\cite{park2021learning,DBLP:conf/iccv/ZhuW21a,RCO,DBLP:conf/aaai/MirzadehFLLMG20}—a prevalent issue in imitation learning~\cite{DBLP:journals/jmlr/RossB10,DBLP:conf/emnlp/LinWCL20}.


In this work, we systematically review existing KD strategies, focusing on two types of distillation (optimization functions and student-generated outputs) to enhance performance. 
Our findings reveal that both enhancement approaches suffer from a common limitation: the fixed distribution of teacher models, which adversely impacts the final student models.
Thus, we propose an online updating method for teachers incorporating student feedback, enabling the teacher model to adapt dynamically to various student sizes. 
Our method facilitates efficient knowledge transfer from the teacher model, accelerating the training process. 
Experimental results across multiple natural language generation datasets demonstrate that our method not only effectively mimics the teacher’s output but also surpasses existing leading methods in model performance. 
Further evaluations involving optimization functions, model sizes, exposure bias, and diverse online module architectures substantiate the robustness and effectiveness of our approach.
Notably, our method achieves superior results compared to state-of-the-art approaches like GKD~\cite{agarwal2023gkd} and MiniLLM~\cite{gu2023minillm}, accomplishing this in just 1/4 of the training time, thereby significantly enhancing efficiency.
Our contributions are:
\begin{itemize}
    \item Our study reveals the phenomenon of difficulty preference in the RKL approach and that changing the fixed teacher distribution is suboptimal (\S \ref{analysis1}).

    \item Our study highlights that the exposure bias problem in KD for auto-regressive language models remains significant when the size gap between teacher and student is large. Moreover, even if this bias is reduced, it may potentially impair the distillation effectiveness (\S \ref{effect_sgo}).

    \item We introduce the idea of online distillation and apply it to various generative language models with sizes ranging from 68M to 1B in the instruction-following setting and outperform the state-of-the-art methods by efficiency and performance (\S \ref{okd}).
\end{itemize}

\section{Related Work}
\paragraph{Supervised Finetuning}
Many works~\cite{alpaca,alpaca-gpt4,xu2023baize,phoenix-2023,wang2023far} employ distilled datasets and curate various data for fine-tuning language models, achieving enhanced performance. 
These methods utilize a hyperscale model, such as GPT-4's~\cite{gpt4} distillation data, to perform high-quality distributional mimicry.
Early works like FLAN 2021~\cite{flan2021} and Super-Natural Instructions~\cite{supernli} are to convert traditional NLP tasks into instruction format through manually defined instruction templates. FLAN-CoT~\cite{cot} and FLAN 2022 \cite{flan_v2} employee Chain-of-Thought training prompts to strengthen the reasoning of the model.
\cite{flan-moe} state that instruction finetuning can be thought of as a ``continual finetuning stage''.
\cite{DBLP:conf/acl/SuSKWHOYSZ023} demonstrate how various text encodings represent different tasks, enabling a single model to accomplish multiple downstream tasks and thereby achieving a more generalized model.
These works on supervised fine-tuning predominantly concentrate on fine-tuning large language models, with minimal focus on using instructional fine-tuning data to develop smaller, more efficient models through student imitation of teacher learning.

\paragraph{Knowledge Distillation}
Knowledge distillation is a popular technique for training compacting pre-trained models from a big teacher network to a small student network. 
It can be applied to many domains like natural language processing~\cite{meta-kd,ding2021understanding,jing23seq}
, image-text retrieval~\cite{rao2023DCD} and image classification~\cite{DBLP:conf/aaai/MirzadehFLLMG20,CRD,DBLP:conf/cvpr/ParkKLC19/RKD}. 
Some works of online learning ~\cite{DML, KDCL,pesf-kd} evaluated on image classification focus on updating all parameters in both networks using labels (hard labels) and feedback information (soft labels) in the training process. 
This approach necessitates updating the parameters of all teachers or relies heavily on data augmentation during training, rendering it less suitable for more complex generative scenarios~\cite{jing22nersurvey,ding2021rejuvenating,ding2022redistributing,xiang2024dkdm}.
Existing studies have primarily focused on simpler distributions (classification tasks), and the recent development of large language models (LMs) underscores an urgent need to explore distillation within generative task scenarios. 
Recent discussions~\cite{wu2024rethinking,gu2023minillm} on Kullback-Leibler Divergence in large LMs distillation have intensified. Much of this work~\cite{wen-etal-2023-f,agarwal2023gkd,kim2024promptkd,ko2024distillm}
aims to mitigate the training-inference mismatch issue, which is exacerbated by the extensive action space inherent in generative task scenarios. 
Although these knowledge distillation methods yield improved results, their reliance on student-generated output during the training process significantly extends training time, which limits their practical utility and shows much lower performance when the size between teacher and student is large. 


\section{Preliminary}\label{revisit}

\subsection{Standard KD for Autoregressive Language Models}

\paragraph{Supervised Finetuning and Forward KD} For a training data source $\mathcal{D}=\left\{\boldsymbol{x}^n, \boldsymbol{y}^n\right\}$, 
the standard instruction tuning of the language model is trained with maximum likelihood estimation, and the training objection can be calculated by Equation \ref{eq:mle}: 
\begin{equation}\label{eq:mle}
\begin{aligned}
\mathcal{L}_{MLE}(\theta) & = \underset{}- \sum_{n=1}^N \sum_{m=1}^M \log \boldsymbol{p}\left(\boldsymbol{y}^n_m \mid \boldsymbol{y}^n_{<m} ; \boldsymbol{x}^n ; \boldsymbol{s}^n; {\theta}\right),
\end{aligned}
\end{equation}
where $\boldsymbol{s}, \boldsymbol{x}$ and $\boldsymbol{y}$ represent the instruction, the input, and the target, respectively. The ${\theta}$ are the parameters to be optimized during the language model training and $M$ is the sequence length of the target.
Forward KD~\cite{DBLP:journals/corr/HintonVD15} often employs a pre-trained teacher network to transfer the teacher's knowledge to a small group of students in the classification task. 
One of the simplest forms is to provide the soft label information by forwarding the teacher's output. The probability prediction of teacher and student model can be defined as: teacher $\boldsymbol{p}(\theta^t)$ and student $\boldsymbol{p}(\theta^s)$, respectively, where $\theta$ is the model parameters and $\boldsymbol{p}(\cdot)$
is the probability prediction.
 For autoregressive language models, KD is a widely used technique where the student is trained to imitate the token-level probability distributions of the teacher. The student $\theta^s$ is trained with the supervised objective $\mathcal{L}_{KL}$ over the target token-level probabilities of the teacher. So the loss measuring the KL-Divergence of teachers and students can be formulated as:

\begin{equation}\label{eq3}
\begin{aligned}
\mathcal{L}_{K L}\left(\boldsymbol{p}(\tau;\theta^s), \boldsymbol{p}(\tau;\theta^t)\right)&=\\
&-\tau^{2} \sum_{n=1}^N\sum_{m=1}^M\boldsymbol{p}_{}(\boldsymbol{y}^n_m|\boldsymbol{y}^n_{<m};\tau;\theta^t)\\
&\cdot\log \frac{\boldsymbol{p}_{}(\boldsymbol{y}^n_m|\boldsymbol{y}^n_{<m};\tau;\theta^t)}{\boldsymbol{p}_{}(\boldsymbol{y}^n_m|\boldsymbol{y}^n_{<m};\tau;\theta^s)},
\end{aligned}
\end{equation}
where $\tau$ is the temperature, which controls how much to rely on the teacher’s soft predictions. 
However, the LLMs require the low-capacity student models to imitate the complex text generation distribution of the teacher models or humans~\cite{ji2023tailoring}, which is hard for Forward KD. 
\begin{figure*}
    \centering
    \vspace{10pt}
\includegraphics[width=0.8\linewidth]{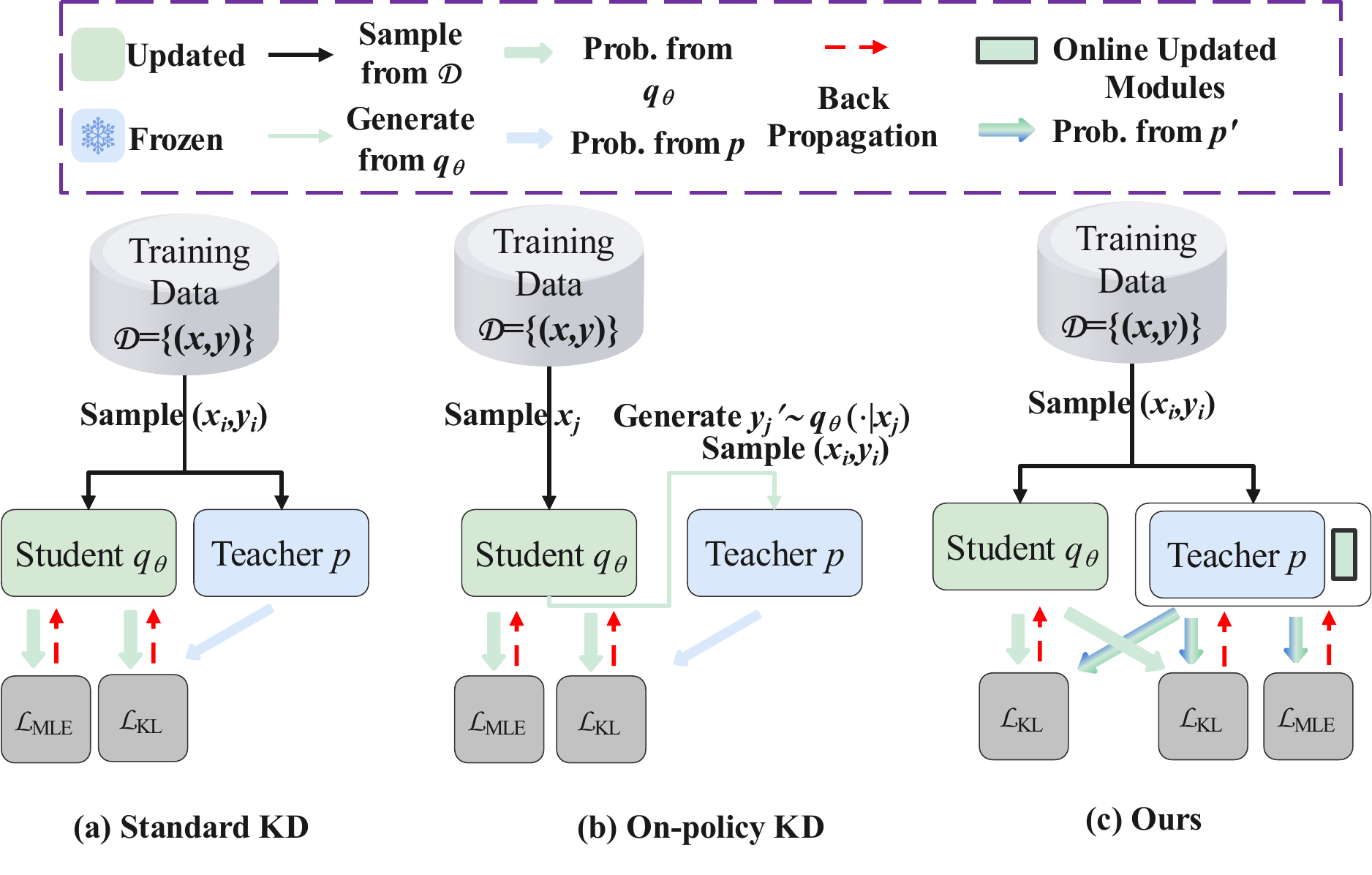}
    \caption{An overview of the KD methods in auto-regressive LMs.}\label{overview}
\vspace{-15pt}
\end{figure*}

\paragraph{Reverse KD} The standard KD methods approximately minimize the Forward KL, as previously stated. 
Reverse KD~\cite{wen-etal-2023-f} ($\mathcal{L}_{K L}\left(\boldsymbol{p}(\tau;\theta^t), \boldsymbol{p}(\tau;\theta^s)\right)$) fundamentally differs from forward KD by interchanging the roles of the teacher and student in the KL Divergence formula, resulting in distinct objectives due to the asymmetry of KL Divergence.
The Reverse KD causes the mode-seeking behavior~\cite{gu2023minillm} in generative modeling, which is crucial to ensure the correctness and faithfulness of text generation. 
Figure \ref{overview} (a) illustrates the training flow of this approach, where the teacher-student samples the same data and the student updates the student network by optimizing the MLE and distillation loss ($\mathcal{L}_{KL}{\left(\boldsymbol{p}(\tau;\theta^s), \boldsymbol{p}(\tau;\theta^t)\right)}$ or $\mathcal{L}_{KL}\left(\boldsymbol{p}(\tau;\theta^t), \boldsymbol{p}(\tau;\theta^s)\right)$). This methodology follows the standard distillation process, whereby students distill knowledge by imitating the teacher's distribution.

\subsection{On-policy KD for Autoregressive Language Models}
As depicted in Figure \ref{overview} (b), the primary distinction between on-policy KD and standard KD lies in the modification of training input data. 
Specifically, in on-policy KD, the teacher's input data is generated by the students during the training process. This modification helps alleviate the training-inference mismatch issue~\cite{agarwal2023gkd,gu2023minillm} but also results in increased training time. At the same time, such methods still suffer from the under-specification problem~\cite{park2021learning,DBLP:conf/iccv/ZhuW21a,RCO,DBLP:conf/aaai/MirzadehFLLMG20}.

\subsection{Evaluation Metrics}
\textbf{ROUGE-L} is an evaluation metric commonly used in assessing the quality of summaries and machine translations,  which 
is suitable for large-scale instruction-following evaluation~\cite{flan_v2,supernli}.  It measures the similarity between a candidate summary/translation and a set of reference summaries/translations by calculating the longest common subsequence (LCS) between them. ROUGE-L~\cite{rouge} considers the structural similarity at the sentence level and is thus suitable for evaluating summaries that capture the main idea of the original text. 

\textbf{TOP 1 Agreement rate} measures the similarity in the TOP 1 predictions between teacher and student models at each position when using teacher-forcing mode. A higher TOP 1 Agreement value indicates that the predictive performance of the student model more closely approximates that of the teacher model~\cite{zhang-etal-2023-towards-understanding}. The TOP 1 Agreement (TA) Rate, calculated over an entire sentence, can be expressed mathematically as follows:
\begin{equation}
    \text{TA} = \frac{1}{N} \sum_{i=1}^N \mathbf{1}(\operatorname{argmax}(p^t_{i}) = \operatorname{argmax}(p^s_{i}))
\end{equation}
where $p^t_{i}$ represents the probability distribution at the $i$-th step for the teacher model, $p^s_{i}$ is that for the student model, and $N$ denotes the sequence length.
In this work, we utilize the token-level difficulty metric, uncertainty coefficient (UNC)~\cite{zhong2024revisiting}, to quantitatively evaluate the variation in the TOP 1 Agreement among different approaches when applied to both complex and simple samples. 

The computational methodology for the UNC metric is specified as follows:
\begin{equation}
    \text {UNC} = \frac{\sum_{k=1,k\neq g_t}^C \operatorname{exp}(z_k^t)}{\sum_{j=1}^C\operatorname{exp}(z_j^t)}
\end{equation}
where $C$ is the vocabulary size, $g_t$ denote the target token/class at $t$-th step, $z_i^t$ represents the logit of $i$-th class in vocabulary.

\textbf{ExAccErr} is the metric for evaluating the exposure bias.
The exposure bias up to $l$ generation steps can be quantified as follows:
\begin{equation}
\operatorname{ExAccErr}(l)=\frac{R(l)-l \epsilon(l)}{l \epsilon(l)} \times 100 \%,
\end{equation}
where $R(l)$ is the accumulated regret of imitating the teacher distribution $p$ at the time step $l$ during the free-run generation, and $\epsilon(l)$ is the average per-step error between $q_\theta$ and $p$ using the oracle context sampled from $p$ as the prefix:
\begin{equation}
R(l)=\sum_{t=1}^T \underset{\substack{\boldsymbol{y}_{<t} \sim q_\theta(\cdot \mid \boldsymbol{x}) \\ y_t \sim p(\cdot \mid \boldsymbol{y}<t}}
{\mathbb{E}} \log \frac{p\left(y_t \mid \boldsymbol{y}_{<t}, 
\boldsymbol{x}\right)}{q_\theta\left(y_t \mid \boldsymbol{y}_{<t}, \boldsymbol{x}\right)}, 
\end{equation}
\begin{equation}
\epsilon(l)=\frac{1}{l} \sum_{t=1}^T \underset{\substack{\boldsymbol{y}_{<t} \sim p(\cdot \mid \boldsymbol{x}) \\ y_t \sim p\left(\cdot \mid \boldsymbol{y}_{<t}, \boldsymbol{x}\right)}}{\mathbb{E}} \log \frac{p\left(y_t \mid \boldsymbol{y}_{<t}, \boldsymbol{x}\right)}{q_\theta\left(y_t \mid \boldsymbol{y}_{<t}, \boldsymbol{x}\right)}.
\end{equation}
\subsection{Experimental Setup}
We compare the widely-used KD methods discussed in Section \ref{revisit}: Forward KD and Reverse KD. We also evaluate recently proposed ImitKD~\cite{DBLP:conf/emnlp/LinWCL20}, and GKD \footnote{We do not combine RL optimization on GKD to make a fair comparison. }variants~\cite{agarwal2023gkd}. 
We also compared our method with the basic fine-tuning approach, namely Supervised Fine-Tuning (SFT).
Our experiments start from student and teacher models with different sizes and architecture, specifically GPT-2 family~\cite{radford2019language} and open-sourced LLaMa ~\cite{llama,zhang2024tinyllama}.
For GPT-2, we use the supervised fine-tuned GPT2-XLarge model on the Dolly 15K as the teacher. For students, we use the GPT-2 base/medium/large checkpoints provided by \cite{gu2023minillm} as the students.\
For LLaMa models, we use supervised fine-tuned LLaMA2-7B/13B trained on Alpaca~\cite{alpaca-gpt4} 52K as the teacher. We use Llama-1.1 B/68M for the student, far smaller than the teacher. 
We report the
average RougeL scores across 5 random seeds for better reproducibility~\cite{rao2022reproducibility}. We evaluate the trained models on 4 instruction-following datasets: Dolly \cite{DatabricksBlog2023DollyV2}: the 500-sample test set we split from the databricks-dolly-15K dataset. Vicuna \cite{vicuna}: The 80 challenging questions used in the Vicuna evaluation. SelfInst~\cite{wang-etal-2023-self-instruct}: A user-oriented instruction-following set with 252 samples. S-NI: The test set of \textsc{Super-NaturalInsrtuctions} \cite{supernli} consisting of 9K samples ranging from 119 tasks.

\subsection{Training Details}\label{experiment_details}
For the results of LLaMA, we follow the experiment setup of \cite{zhong2024revisiting} for better performance.
For models with 1B parameters, we perform a hyperparameter search over the learning rate in $\{$5e-4, 1e-4, 5e-5$\}$ and the batch sizes in $\{8,16,32\}$, with no weight decay and a learning rate with linear decay and linear warmup for 3\% of the total training steps. For models that have more than 1B parameters, we search for the learning rate in $\{$5e-5, 1e-5, 5e-6$\}$, the batch sizes of 8, and train these models for 3 epochs.  
We use a maximum sequence length of 512 for GPT and 2048 for LLaMA, truncating samples where necessary. 
During training, we use the DeepSpeed library~\cite{deepspeed} and ZeRO~\cite{zero} optimizer to allow for large-scale model finetuning.  For analytical experiments, we follow \cite{gu2023minillm} and use the provided code with the same method, removing a language modeling loss for fair comparison with each baseline.
We first construct the training data from dolly-15k~\cite{DatabricksBlog2023DollyV2}, wherein we randomly select 14K samples for training and equally leave 500 samples for validation and testing, respectively. We train all models for analysis in 20 epochs. All experiments were performed on an 8$\times$ A800-80G server.

\section{Understanding Knowledge Distillation in Autoregressive Language Models}
\subsection{Two Aspects of Improving KD in Autoregressive LMs}

\paragraph{Optimization Function}
Reverse KD (RKD)~\cite{gu2023minillm,wen-etal-2023-f} has been successfully applied in auto-regressive language models for knowledge distillation. To analyze the source of its success, we divided the Dolly validation set by difficulty~\cite{zhong2024revisiting} and systematically compared the standard deviation (STD), the KL divergence, and the agreement of the TOP 1 predictions~\cite{zhang-etal-2023-towards-understanding}. Specific explanations of the metrics can be found in the Appendix \ref{metric}.
We compared the performance of Forward KD (FKD) and JSD~\cite{wen-etal-2023-f} and RKD at different temperatures.

\paragraph{Student-Generated Output}\label{distribution_sim}
Exposure Bias~\cite{arora-etal-2022-exposure} refers to the distributional mismatch between sentences observed during training and those generated during inference. 
This bias can lead to deviations in the tokens produced during inference from those encountered in training, culminating in cumulative errors throughout the generation process. 
A notable advancement in on-policy KD is the integration of student-generated output. 
This approach has been demonstrated to reduce mismatch metrics~\cite{arora-etal-2022-exposure}.
Our analysis examines potential issues with existing on-policy KD strategies by assessing the exposure bias using ExAccErr~\cite{arora-etal-2022-exposure} and the model's adaptive response during validation using the validation loss and RougeL~\cite{rouge} metric.

\subsection{Effect of the Optimization Function}\label{analysis1}


From an optimization function perspective, using Forward/Reverse KL can lead to specific behaviors in the student model. For the sake of simplicity of the next description, we define that $P$
 is the fixed teacher's distribution and 
$Q$ is the adaptable student distribution.
Forward KL and Reverse Divergence are defined as follows: 
\begin{equation}
\begin{aligned}
D_{K L}(P \| Q)=\int P(x) \log \frac{P(x)}{Q(x)} d x, \\
D_{K L}(Q \| P)=\int Q(x) \log \frac{Q(x)}{P(x)} d x
\end{aligned}
\end{equation}

\textbf{Proof. Considering the converge condition for FKL.} 
To minimize \( D_{KL}(P \parallel Q) \) with respect to a parameterized \( q(x) \) modeled with parameters \( \theta \), the derivative with respect to \( \theta \) is set to zero. The derivation is as follows:

\begin{equation}
\begin{aligned}
\frac{\partial}{\partial \theta} D_{KL}(P \parallel Q) &= \frac{\partial}{\partial \theta} \int p(x) \log \frac{p(x)}{q(x)} \, dx \\
&= -\int p(x) \frac{\partial}{\partial \theta} \log q(x) \, dx \\
&= -\int p(x) \frac{1}{q(x)} \frac{\partial q(x)}{\partial \theta} \, dx
\end{aligned}
\end{equation}

To achieve minimization, the following conditions must hold:

\begin{equation}
0 = -\int p(x) \frac{1}{q(x)} \frac{\partial q(x)}{\partial \theta} \, dx
\end{equation}

This condition implies the expectation of the logarithmic derivative of \( q(x) \) with respect to \( \theta \) under the distribution \( P \) should be zero:

\begin{equation}
\mathbb{E}_{x \sim P}\left[\frac{\partial}{\partial \theta} \log q(x)\right] = 0
\end{equation}

This can be interpreted as maximizing the expected log-likelihood of the distribution \( Q \) under \( P \), making \( q(x) \) as close as possible to \( p(x) \) by the parameterized model.

\textbf{Proof. Considering the converge condition for FKL.} 
To minimize \( D_{KL}(Q \parallel P) \) with respect to a parameterized \( q(x) \) modeled with parameters \( \theta \), the derivative with respect to \( \theta \) is set to zero. Using the fact that $\log \frac{q(x)}{p(x)}=\log q(x)-\log p(x)$ and that $p(x)$ does not depend on $\theta$, the differentiation proceeds as:

\begin{equation}
\begin{aligned}
\frac{\partial}{\partial \theta} D_{KL}(Q \parallel P) &= \frac{\partial}{\partial \theta} \int q(x) \log \frac{q(x)}{p(x)} \, dx \\
&= \int \frac{\partial q(x)}{\partial \theta} \log \frac{q(x)}{p(x)} \, dx \\ 
&+ \int q(x) \frac{1}{q(x)} \frac{\partial q(x)}{\partial \theta} \, dx \\
\end{aligned}
\end{equation}

The second integral is the derivative of the total probability (which must be 1). Thus, it equals zero. Therefore, we simplify further:
\begin{equation}
\begin{aligned}
\frac{\partial}{\partial \theta} D_{KL}(Q \parallel P) 
&= \int \frac{\partial q(x)}{\partial \theta} \log \frac{q(x)}{p(x)} \, dx + \int \frac{\partial q(x)}{\partial \theta} \, dx \\
&= \int \frac{\partial q(x)}{\partial \theta} \log \frac{q(x)}{p(x)} \, dx
\end{aligned}
\end{equation}
Setting this derivative to zero for optimization gives:
\begin{equation}
\int \frac{\partial q(x)}{\partial \theta} \log \frac{q(x)}{p(x)} \, dx = 0
\end{equation}

This condition implies that the weighted average (weighted by \(\frac{\partial q(x)}{\partial \theta}\)) of the logarithm of the ratio \(\frac{q(x)}{p(x)}\) must be zero. This can be interpreted as minimizing the information gain (or loss) when moving from distribution \(Q\) to \(P\), effectively making \(q(x)\) as similar as possible to \(p(x)\) by adjusting the parameters \( \theta \). This is especially important in scenarios such as variational inference, where it helps in tightly concentrating the variational distribution around the true distribution's mass.

\textbf{Principle 1: }
Forward KL Divergence emphasizes reducing instances where the student model 
$Q(x)$ predicts probabilities that are significantly lower than those of the teacher model $P(x)$ for events where $P(x)$ is high. When $Q(x)$ is substantially less than $log\frac{P(x)}{Q(x)}$ becomes large, thus increasing the overall KL divergence and imposing a significant penalty on the model. This leads to mean-seeking. 

\textbf{Principle 2: }When the student model's probability 
$Q(x)$ significantly exceeds the corresponding probability $P(x)$ in the teacher model. The term $log\frac{Q(x)}{P(x)}$ becomes large (positive). Thus, the entire expression also becomes large, indicating a significant penalty. 
This situation pushes the student model to reduce these probabilities and leads to mode-seeking. 

\textbf{Principle 3: }
In practice, a straightforward method to alter the data focus of the optimization process by controlling the distribution of teachers 
$P(x)$ by introducing a hyperparameter 
$T$ greater than 1 to reduce the denominator of RKL in the original equation.
This leads to a higher occurrence of {Principle 2} cases during training. This adjustment decreases the probability value assigned by the student model to this subset of the data. Essentially, by incorporating more difficult samples relative to the teacher model, the optimization process focuses on a broader range of data.



\begin{figure*}[t]
\centering
\subfigure[GPT-2 XLarge to GPT-2 Base]
{
    \begin{minipage}[t]{1.6\columnwidth}
    \centering
\includegraphics[width=\linewidth]{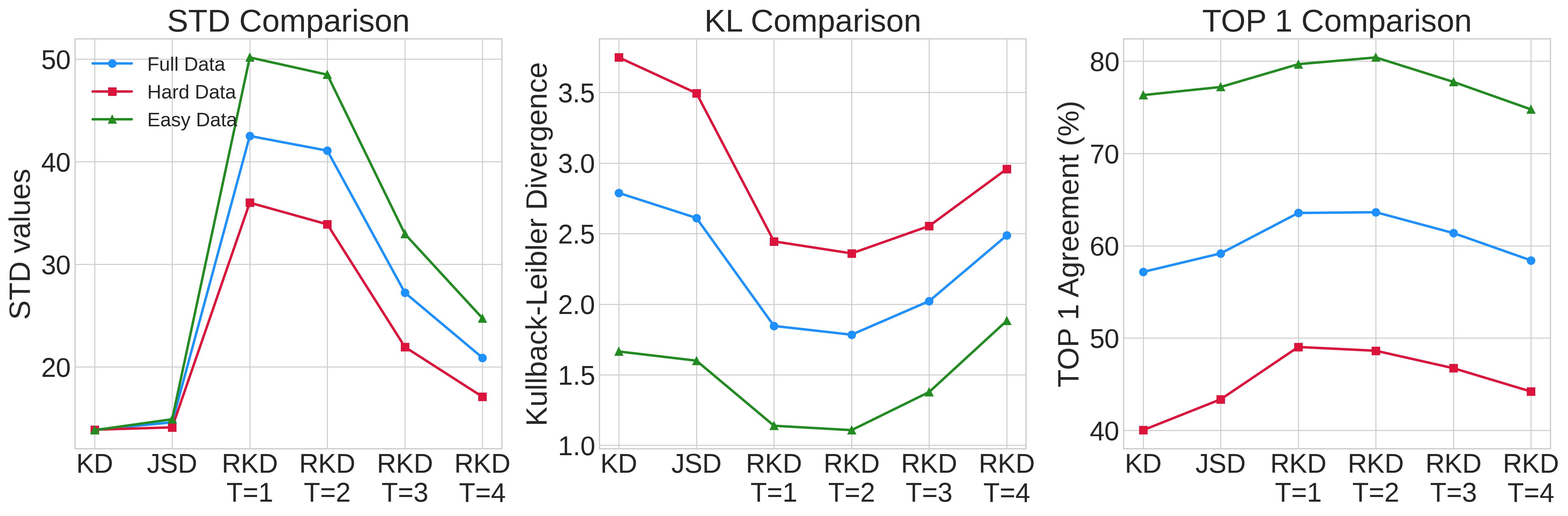}
\end{minipage}    
}
\\
\subfigure[GPT-2 XLarge to GPT-2 Large]
{
\begin{minipage}[t]{1.6\columnwidth}
    \centering
\includegraphics[width=\linewidth]{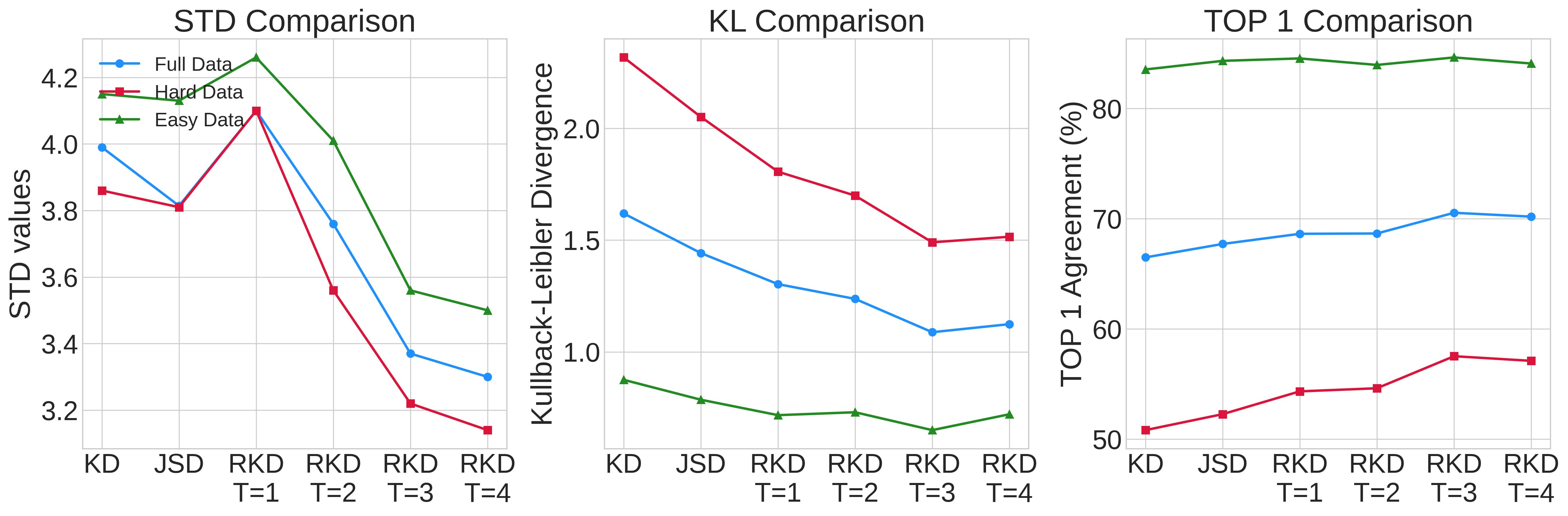}
\end{minipage}  
}
\vspace{-0.8em}
    \caption{Comparison of metrics such as standard deviation for logits (STD), teacher-student prediction similarity (KL), and TOP 1 prediction agreement for different optimization functions. 
    }\label{kl_compare}
\end{figure*}

\textbf{Findings 1}: Forward KL generally optimizes all tokens uniformly, whereas Reverse KL focuses on more challenging tokens. 
As indicated by Principle 1, KD demonstrates similar standard deviations (STDs) for logits across data of varying complexities, suggesting a mean-seeking behavior as illustrated in Figure~\ref{kl_compare}. According to Principle 2, RKL specifically targets the more challenging segments of data (those with lower probability $P(x)$). This results in a more gradual decrease in KL divergence for simpler data and a sharper decrease for more complex data. Such a pattern is also evident in the TOP 1 predictions and STDs, indicating that RKL intervention leads to training that concentrates on difficult samples.

\textbf{Findings 2}: Adjusting the temperature and utilizing JSD can shift the focus of model training based on sample difficulty.
We analyzed the impact of different temperatures in RKD. The results, consistent with Principle 3, show that increasing the temperature (e.g., T=2) enhances model effectiveness, as demonstrated by a significant reduction in STD and KL for challenging data alongside a modest reduction for simple data. Additionally, TOP 1 accuracy for simple data improves, while accuracy for complex data declines, suggesting a shift in focus towards simpler samples when the temperature is adjusted. Conversely, excessively high temperatures (e.g., T=4) diminish model performance, in line with Principle 2, by rendering the optimization of overly simplistic samples less effective.

\textbf{Discussion}:
In the NLG scenario, since the generation sequence is more complex than the classification (satisfying Principle 2), Reverse KD may be more effective than Forward KD.
However, the divergence metric optimizes the objective by focusing on certain parts of the teacher distribution but does not adequately capture the overall information across different and complex NLG tasks. If optimization for different tasks is considered, the fixed teacher distribution needs to be modified.



\subsection{Effect of the Distribution on the Student-Generated Output}\label{effect_sgo}

\begin{table*}[t]
\caption{
The excess error caused by the training-inference discrepancy (ExAccErr) results of different KD methods. 
The ExAccErr values closer to 0 mean the method introduces less exposure bias. Forward KD suffers exposed bias, and other on-policy KD methods alleviate this issue.
}\label{ece}
\centering
\scalebox{1.0}
{
\begin{tabular}{rcccccc}
\toprule 
\multirow{2}{*}{Method}&\textbf{Baseline}&\multicolumn{2}{c}{\textbf{Standard KD}}&\multicolumn{3}{c}{\textbf{On Policy}}
\\
\cmidrule(lr){2-2} \cmidrule(lr){3-4} 
\cmidrule(lr){5-7}
& SFT & \begin{tabular}[c]{@{}l@{}}Forward \\\end{tabular} & \begin{tabular}[c]{@{}l@{}}Reverse\\\end{tabular} & ImitKD&GKD  & \begin{tabular}[c]{@{}l@{}}MiniLLM\end{tabular} \\ 
\midrule
NLG AVG.  &16.1&16.4&18.1&15.6&19.6&19.7\\
ExAccErr & 22.1   & 34.6                                                     & 16.3                                                    &11.8& -6.6   & 0.8                                                                                                      \\ \bottomrule
\end{tabular}
}
\end{table*}


\begin{figure}
\centering
\begin{minipage}[t]{0.45\linewidth}
\centering\includegraphics[width=1\linewidth]{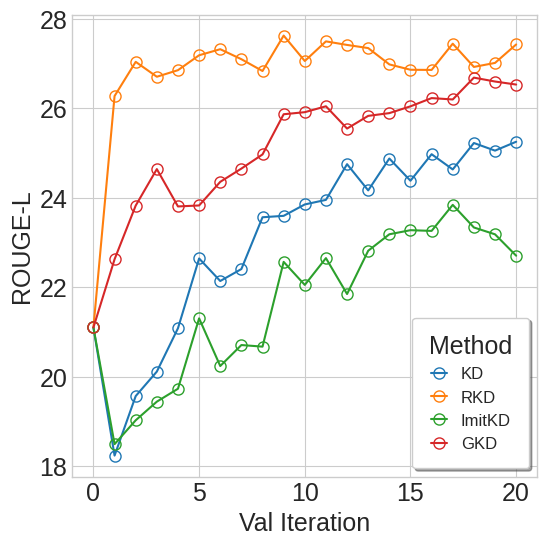}
\vspace{-1.5em}
    \caption{ROUGE-L scores for the validation set across the different methods.}
    \label{rouge}
\end{minipage}
\hspace{0.7em}
\begin{minipage}[t]{0.45\linewidth}
    \centering
\includegraphics[width=1\linewidth]{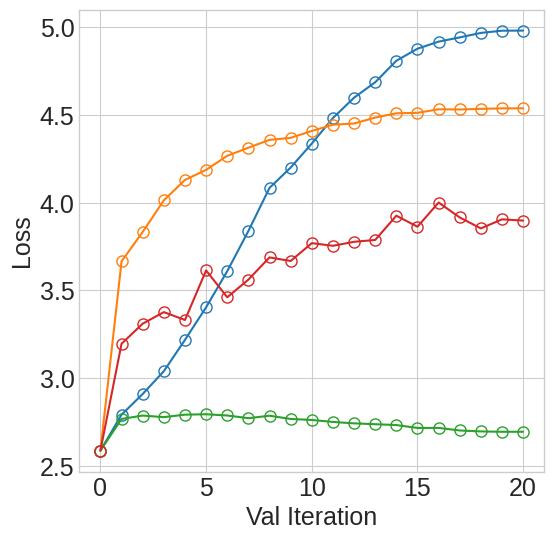}
\vspace{-1.5em}
    \caption{Plot of validation loss values across each validation iteration. 
    }
    \label{val_loss}
\end{minipage}
\hspace{0.7em}
\end{figure}

In Figure \ref{rouge} and Figure \ref{val_loss}, we compare the results of various knowledge distillation methods, spanning standard to state-of-the-art, on the RogueL and loss metric with the validation set across training epochs. Results are tested on GPT-2 XLarge/Base.

\textbf{Findings 1}: On-policy methods only mitigate exposure bias on certain metrics. 
As outlined in Section \ref{distribution_sim}, we use ExAccErr as a metric to assess the extent of training-inference mismatch, comparing commonly used methods as shown in Table \ref{ece}. These results support the hypothesis~\cite{gu2023minillm} that Forward KD is unsuitable for distilling large language models and tends to result in a training-inference mismatch. Conversely, methods like reverse KD and on-policy KD have shown effectiveness in mitigating this issue, showing the lower ExAccErr.
On-policy method ImitKD employs student-generated tokens during training, significantly reducing the training-inference mismatch, as indicated by the lowest loss (green line) in Figure \ref{val_loss}. 
The more recent GKD method not only effectively reduces this mismatch, as demonstrated by lower validation set losses, but also achieves the highest RougeL score. 
However, in Figure \ref{val_loss}, the validation set loss for all methods increased during training, suggesting a possible mismatch between the training and inference phases. 
This same trend illustrates that existing methods still have serious exposure bias problems.

\textbf{Finding 2}: Using student-generated outputs when the student is poor can impair distillation results.
As shown in Figure \ref{rouge} for ImitKD, integrating student-generated and training set distributions initially results in lower validation scores, indicative of early adjustment difficulties to align with the student distribution, as seen from epoch 0 to 1. GKD and RKD, in contrast, quickly achieve higher scores early in training, leading to superior final RougeL scores. However, as shown in Figure \ref{rouge} and Table \ref{ece}, the low RougeL scores and NLG average scores for ImitKD also reflect a decrease in distillation effectiveness caused by the significant divergence between the student-generated distribution and the teacher’s previously learned distribution.
This indicates that the disparity in size between student and teacher models plays a crucial role in the effectiveness of student-generated outputs on the final performance.

\textbf{Discussion}:
The core principle of on-policy KD involves modifying the input data distribution by allowing students to generate their own input data. 
This strategy enhances the diversity of the training data and mitigates overfitting. 
However, this self-generated data distribution might be unfamiliar to the teachers or prove suboptimal, particularly when there is a notable disparity in model sizes. This leads to the difficulty of distribution transfer.
Such discrepancies are evident in the performance of ImitKD in terms of RougeL and GKD in the loss metric, which could potentially compromise the effectiveness of the final distillation process.
 
\begin{table*}[t!]
\caption{GPT2 results for natural language generation tasks. 
* represents the results copy from \cite{gu2023minillm}, which used the extra and large pre-training dataset openwebtext. Our results usually outperform MiniLLM, and the training time is shorter.
We report the average R-L scores across 5 random seeds. 
}\label{tab:gpt}
\centering
\begin{tabular}{{lllcccccl}}
\toprule
\textbf{Model}&\#\textbf{Params}&\multirow{1}{*}{\textbf{Method}} & \textbf{Time}&\multicolumn{1}{c}{\textbf{DollyEval}}  &  
\multicolumn{1}{c}{\textbf{VicunaEval}}&\textbf{SelfInst}&\textbf{S-NI} &{\textbf{AVG.}}          \\ \midrule
\multirow{24}{*}{\textbf{GPT-2}} &\textbf{1.5B}&\textbf{Teacher}&
          -&                   27.6&	16.3&	14.3&	27.6&	21.5
 \\
          \cmidrule(l){2-9}
&\multirow{8}{*}{\textbf{120M}}&\textbf{Student}
          & &&& \\
&&\quad+SFT &-& 23.3&                  14.7&10.0&16.3&16.1\\
&&\quad+Forward KD&0.6& 23.7&	14.8&	10.2&	16.8&	16.4     
         \\
&&\quad +Reverse KD&0.6&25.0	&15.9&	11.3&	20.3&	18.1 
\\
&&\quad +ImitKD&4.8&20.7&	14.7&	10.0&	17.2&	15.6
\\
&&\quad +GKD&3.6&24.9	&16.1&	13.1&	24.1&	19.6 \\
&&\quad +MiniLLM*&5.3&24.6&	16.9&	13.2&	25.3&	20.0\\
&&\quad+OKD &1.5& 25.7	&14.7&	12.8&	26.4&	19.9 (+3.8)
\\
          \cmidrule(l){2-9}
&\multirow{8}{*}{\textbf{340M}}&\textbf{Student}
          & &&& \\
&&\quad+SFT &-& 25.5&	16.0&	13.0&	25.1&	19.9
\\
&&\quad+Forward KD&1.4&25.0&15.4	&12.0&	23.7&	19.0
         \\
&&\quad +Reverse KD&1.4&26.2&	17.1&	14.4&	26.3&	21.0
\\
&&\quad +ImitKD&7.0&21.6	&14.7&	10.8&	17.9&	16.3\\
&&\quad +GKD&8.5&26.2&	16.5&	14.4&	29.1&	21.6\\
&&\quad +MiniLLM*&9.0&25.4&	17.7&	15.6&	27.4&	21.5\\
&&\quad+OKD&2.6& 27.5&	17.0&	15.5&	30.9&	22.7 (+2.8)
\\
          \cmidrule(l){2-9}
&\multirow{8}{*}{\textbf{760M}}&\textbf{Student}
          & &&& \\
&&\quad+SFT &-& 25.4&	16.1&	12.4&	21.5	&18.9
\\
&&\quad+Forward KD&2.6&26.3&	16.1&	12.4&	22.7&	19.4
         \\
&&\quad +Reverse KD&2.7&26.2&	16.6&	15.1&	26.7&	21.2
\\
&&\quad +ImitKD&14.6&24.1	&15.8	&12.8	&23.7	&19.1
\\
&&\quad +GKD&10.5&26.8&	16.7&	14.4&	28.5&	21.6\\
&&\quad +MiniLLM*&16.4&26.4&	18.3&	15.9&	29.3&	22.5\\
&&\quad+OKD & 4.2&26.5&	15.9&	16.8&	32.2&	22.9 (+4.0)
\\ \bottomrule
\end{tabular}
\end{table*}

\section{Alternative Method for Knowledge Distillation in Autoregressive LMs}\label{okd}

\subsection{Online Knowledge Distillation}
Building on the previous discussion, 
we aim to diversify the training data distribution by altering the fixed distribution of teachers introducing students' feedback.
As illustrated in Figure \ref{overview} (c), it diverges from previous approaches by using raw input data and modifying the teacher’s output distribution through student-teacher feedback during the learning process. 
Smaller models with limited capabilities should closely mimic the teacher's behavior. 
As teachers adaptively change according to the distribution of students, a single forward KL divergence is sufficient for the optimization function. Conversely, larger models need to adapt more flexibly to the small model without being overly influenced by its learning patterns, necessitating the addition of a regularization term (MLE).
The core of online learning is retraining teacher networks, so it is desirable to participate in training a teacher network with only a few parameters. To reduce the over-consumption of the fully trained teacher network, we propose to adopt a standard LoRA module~\cite{lora} as the online module with updating parameters of this LoRA module while the original parameters of the teacher network are fixed.
{The LoRA injects trainable low-rank matrices into transformer layers to approximate the weight updates.
}
For a pre-trained weight matrix
$\boldsymbol{W}\in \mathbb{R}^{d \times k}$, LoRA decomposes $\boldsymbol{W} +\boldsymbol{\Delta W} =
 \boldsymbol{W} +\boldsymbol{W}_\text{down}\boldsymbol{W}_\text{up}$.
For the input $\boldsymbol{x}$ to linear projection in multi-head attention, LoRA utilizes a down-projection weight matrix $\boldsymbol{W}_{\text {down }} \in \mathbb{R}^{d \times r}$ and an up-projection function with weight matrix $\boldsymbol{W}_{\text {up }} \in \mathbb{R}^{r \times k}$ to update to the query and value projection matrice ($\boldsymbol{W}_q,\boldsymbol{W}_v$) in the multi-head attention sub-layer.
{With such down-sampling and up-sampling operations, the feature dimensions of output $h$ by the module are consistent with the dimensions of input $h$.
The output feature $h$ of the LoRA module is as follows:}

\begin{equation}\label{eq6}
\boldsymbol{h} \leftarrow \boldsymbol{h}+s\cdot\boldsymbol{x} \boldsymbol{W}_{\text {down }} \boldsymbol{W}_{\text {up }}
\end{equation}
where $s\geq 1$ is a tunable scalar hyperparameter.
Formally, the training loss of the student and teacher network can be formulated as follows:
\begin{equation}\label{t_loss}
\mathcal{L}_{t}=\alpha\mathcal{L}_{K L}\left(\boldsymbol{p}(\theta^{ta}),
    \boldsymbol{p}(\theta^s)\right)+(1-\alpha)\mathcal{L}_{MLE}(\theta^{ta}), 
\end{equation}
\begin{equation}\label{s_loss}
\mathcal{L}_{s}=\mathcal{L}_{K L}\left(\boldsymbol{p}(\theta^s), \boldsymbol{p}(\theta^{ta})\right),
\end{equation}

where $\theta^{ta}$ is the parameter of the online module of the teacher network needed to update.

\begin{table*}[t!]
\caption{Results for natural language generation tasks on LLaMA-2 family with larger model size difference. GKD does not perform well when the model size gap is too large. 
}\label{tab:main}
\centering
\begin{tabular}{{lllccccl}}\toprule
\textbf{Model}&\#\textbf{Params}&{\textbf{Method}} & \textbf{DollyEval}  &  
{\textbf{VicunaEval}}&\textbf{SelfInst}&\textbf{S-NI}\ &{\textbf{AVG.}}       
\\ \midrule
\multirow{22}{*}{\textbf{LLaMA-2}} &\textbf{7B}&\textbf{Teacher}
          &                   23.3 & 23.1&18.7&34.7&25.0 \\
          \cmidrule(l){2-8}
&\multirow{6}{*}{\textbf{1.1B}}&\textbf{Student}
          &   &	&	&	&               \\
&&\quad+SFT & 16.3&	18.2&	15.2&	25.5&	18.8
         \\
&&\quad+Forward KD& 18.0&	19.2&	14.8&	27.5&	19.9 (+1.1)\\
&&\quad +Reverse KD&20.8& 	22.2&	16.3&	32.2&	22.9 (+4.1)\\
&&\quad +GKD&20.9&	22.4&	16.2&	32.4&23.0 (+4.2)\\
&&\quad+OKD & 22.2&	22.9&	17.8&	38.0&	25.2 (+6.4)\\
\cmidrule(l){2-8}
 &\textbf{13B}&\textbf{Teacher}
          & 23.0&	25.1&	18.1&	38.9&	26.3
\\
\cmidrule(l){2-8}
&\multirow{6}{*}{\textbf{68M}}&\textbf{Student}
          & &   &		&	&                    \\
&&\quad+SFT & 9.3&	12.2&	6.3&	7.3&	8.8
         \\
&&\quad+Forward KD&
9.2&	12.2&	6.2&	7.3&	8.7 (-0.1)
         \\
&&\quad +Reverse KD&
9.3&	17.0	&6.1&	4.8&	9.3 (+0.6)
\\
&&\quad +GKD&9.1	&13.1	&6.2&	6.6&	8.8 (+0.1)
\\
&&\quad+OKD & 11.9&	15.1&	7.9	&10.1&	11.3 (+2.6)
\\ \cmidrule(l){2-8}
&\multirow{6}{*}{\textbf{1.1B}}&\textbf{Student}
          & &   &	&	&	                    \\
&&\quad+SFT & 16.3&	18.2&	15.2&	25.5&	18.8
         \\
&&\quad+Forward KD&18.1&	17.4&	14.5&	29.0&	19.8 (+1.0)
         \\
&&\quad +Reverse KD&20.7&	22.4&	15.8&	35.6&	23.6 (+4.8)
\\
&&\quad +GKD&20.5&	22.9&	15.9&	36.0&	23.8 (+5.0)
\\
&&\quad+OKD & 23.1	&21.7&	18.3&	38.9&	25.5 (+6.7)
\\ \bottomrule
\end{tabular}
\end{table*}

\begin{figure*}[t]
\centering
\subfigure[GPT-2 XLarge to GPT-2 Base]
{
    \begin{minipage}[t]{1.6\columnwidth}
    \centering
\includegraphics[width=\linewidth]{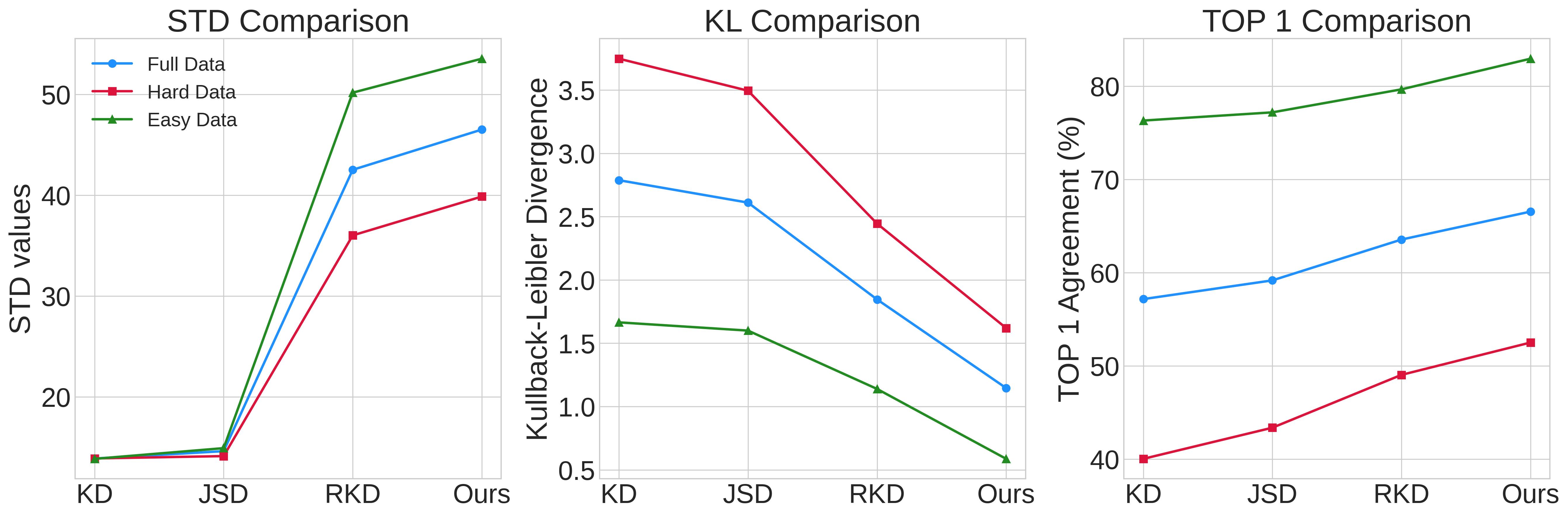}
\end{minipage}    
}
\\
\subfigure[GPT-2 XLarge to GPT-2 Large]
{
\begin{minipage}[t]{1.6\columnwidth}
    \centering
\includegraphics[width=\linewidth]{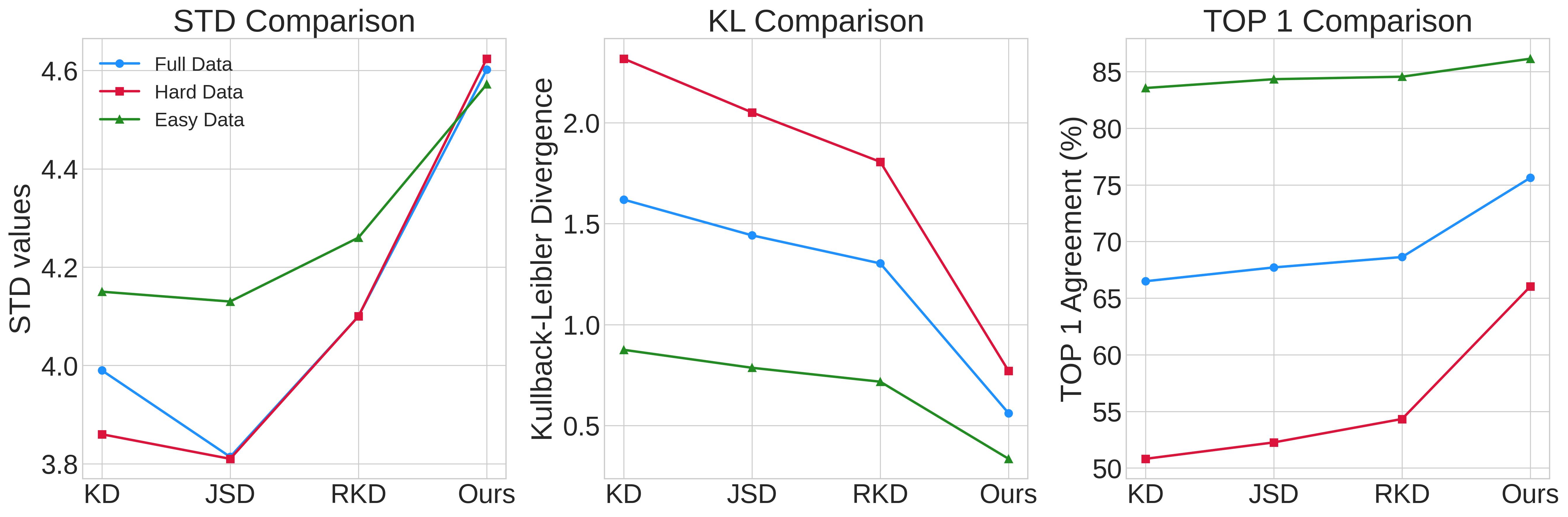}
\end{minipage}  
}
\vspace{-0.8em}
    \caption{Comparison of metrics such as standard deviation for logits (STD), teacher-student prediction similarity (KL), and TOP 1 prediction agreement for different optimization functions.
    Our method achieves better results across samples of varying difficulties and teacher-student combinations.
    }\label{kl_compare_our}
\end{figure*}

\begin{table*}[t]
\caption{
The excess error caused by the training-inference discrepancy (ExAccErr) results of different KD methods.  Existing on-policy methods still have higher exposure bias than the Standard KD in some settings, where our method performs best in this setting.
}\label{tab:ECE_large}
\centering
\begin{tabular}{rcccccc}
\toprule 
\multirow{2}{*}{Method}&\textbf{Baseline}&\multicolumn{2}{c}{\textbf{Standard KD}}&\multicolumn{2}{c}{\textbf{On Policy}}&\multirow{2}{*}{Ours}
\\
\cmidrule(lr){2-2} \cmidrule(lr){3-4} 
\cmidrule(lr){5-6}& SFT & \begin{tabular}[c]{@{}l@{}}Forward \\\end{tabular} & 
\begin{tabular}[c]{@{}l@{}}Reverse\\\end{tabular} & GKD &MiniLLM  \\ 
\midrule
NLG AVG.  &18.9&19.4&21.2&21.6&22.5&22.9\\
ExAccErr & 11.6   & 5.1           & 4.6              & -6.6   & -5.3     &-4.2                \\ \bottomrule
\end{tabular}
\makeatletter\def\@captype{table}\makeatother
\end{table*}



\subsection{Overall Learning Strategy}
We use Algorithm 1 to show the flow of the whole method more directly. 
Specifically, we use the existing labeled data sources, the student model (full parameters), and the teacher model (tuning adapter only) to be fine-tuned as inputs. The whole optimization process ends up with a trained student model. 
In the entire parameter optimization process, we first get the training data samples and then calculate the probability distributions of the student and teacher models on these input samples separately. 
After that, the two optimization objectives mentioned in Equations \ref{t_loss} and \ref{s_loss} are used to update the parameters of the teacher model and the student model until they are trained to the maximum step.
\begin{algorithm}
\caption{OKD}
\textbf{Input:} teacher model $\theta^{ta}$, student model $\theta^{s}$, data source $\mathcal{D} = \{x^n, y^n\}$, training step $K$, learning rate $\eta$ \\
\textbf{Output:} Return the student Model $\theta^s$
\begin{algorithmic}[1]
\For{each step $k = 1, \dots, K$}
    \State Sample a request and a response $(x,y)$ from $\mathcal{D}$
    \State Calculate the probability distributions $p(\theta^{ta})$ and $p(\theta^{s})$ for the given sample
    \State Update $\theta^{ta} \leftarrow \theta^{ta} - \eta \nabla_{\theta^{ta}} \mathcal{L}_t$
    \State Update $\theta^s \leftarrow \theta^s - \eta \nabla_{\theta^s} \mathcal{L}_s$
\EndFor
\State \Return $\theta^s$
\end{algorithmic}
\end{algorithm}
\subsection{Main Results}
We conducted a comparison of established knowledge distillation methods: Forward KD, Reverse KD, and recently proposed on-policy KD methods (ImitKD, GKD, and MiniLLM) in Table \ref{tab:main}.  
For this analysis, the GPT-2 series was used to evaluate the methods, and we referenced results from the original MiniLLM paper~\cite{gu2023minillm} for benchmarking. 
We report generation results and training times for multiple datasets.
Notably, while the state-of-the-art method MiniLLM leverages additional training data, it also introduces inefficiencies associated with student-generated output similar to those observed in GKD. 
Our method exhibits superior performance across the GPT series, outperforming the state-of-the-art methods MiniLLM and GKD in most cases, such as in the transition from 1.5B to 340M models (22.7 vs. 21.6) and achieving closely comparable results in others. 
Our method, not based on an on-policy scheme, significantly accelerates training times compared to on-policy methods, particularly at larger student sizes. For instance, our results with a 760M student model demonstrate that our method achieves a fourfold speedup in training time compared to MiniLLM.





\subsection{Analysis of OKD}

\paragraph{Comparing Optimization Function}
We compared existing optimization functions to demonstrate that combining our methods can achieve better optimization results on Figure \ref{kl_compare_our}. 
The results of two distillation combinations indicate that predictions for different difficulty levels using our methods exhibit a consistent trend: uncertainty decreases (as evidenced by larger standard deviations), the KL divergence decreases, and TOP 1 agreement improves.  
Akin to the phenomenon observed with Reverse KD, our method, combined with Forward KD, tends to focus on challenging samples. This suggests that our approach enhances student learning by modifying the teacher's distribution.
Table \ref{tab:lossfuction_abs} displays the results of employing various optimization functions during the distillation of GPT-2 XLarge to GPT-2 Large. These results indicate that incorporating our online module consistently enhances the performance of the corresponding functions. This improvement across all metrics validates the effectiveness of our approach in overcoming the limitation that specific distillations tend to mimic only certain aspects of a fixed teacher's distribution. 
\paragraph{Comparing Results with Greater Differences in Model Sizes}
To assess knowledge distillation effectiveness in models with large size gaps, we employed LLaMA models as teachers with sizes ranging from 7 to 13 billion parameters, distilling into students at 1 billion and 68 million parameter levels. 
Owing to the underperformance of MiniLLM in our specific settings (lacking pretrain data), it was omitted from our analysis. Instead, we use the state-of-the-art method GKD as a robust baseline. 
Table \ref{tab:main} presents the comparative results, illustrating that our method, OKD, surpasses existing state-of-the-art methods across various teacher-student combinations, particularly in scenarios with significant model size discrepancies. Traditional methods, such as the on-policy GKD, perform suboptimally when faced with large disparities in model size, for example, distilling from a 13 billion parameter model to a 68 million parameter model.
Our approach consistently achieves superior generation results, especially in cases where the gap in model capacity is most pronounced.

\paragraph{Comparing Exposure Bias}
Additionally, we compare the ExAccErr of these methods in Table \ref{tab:ECE_large}, demonstrating our method's significant mitigation of exposure bias within this metric.
As discussed in the preceding analysis Section \ref{effect_sgo}, the on-policy approach shows a reduction in this indicator compared to SFT. 
However, when the difference between teacher and student model sizes is minimal (GPT-2 XLarge to GPT-2 Large), these on-policy methods exhibit a higher exposure bias than Standard KD. 
This observation aligns with findings from the previous validation Section \ref{effect_sgo}, which indicates that exposure bias persists in current KD methodologies.



\begin{table}
\caption{Ablation on the optimization function.
}\label{tab:lossfuction_abs}
\centering
\begin{tabular}{{ll}}
\toprule
\multirow{1}{*}{Method} &{NLG AVG.}          \\ \midrule
Forward  &	19.4
         \\
         \quad+OKD &	22.9 (+3.5)\\
         \cdashline{1-2}
Reverse  &	21.2\\
\quad +OKD &	22.6 (+1.4)\\
\cdashline{0-1}
JSD  &	19.7\\
\quad +OKD &	22.8 (+3.1)
\\ \bottomrule
\end{tabular}
\makeatletter\def\@captype{table}\makeatother
\end{table}

\paragraph{Online Modules Architecture}

\begin{table}
\caption{Ablation on different structures of the online module.
}\label{tab:module_abs}
\centering
\begin{tabular}{{ll}}
\toprule
\multirow{1}{*}{Method} &{NLG AVG.}          \\ \midrule
Teacher
          &                   21.5
 \\
Student
          & \\
\quad+Forward KD &	19.4
         \\
\quad +OKD (P-Tuning)&	21.2 (+1.8)
\\
\quad +OKD (Prefix)&	20.7 (+1.3)
\\
\gray\quad+OKD (LoRA)&	22.9 (+3.5)
\\ \bottomrule
\end{tabular}
\makeatletter\def\@captype{table}\makeatother
\end{table}

The core of our approach is the online module, adaptable to various parameter-efficient fine-tuning methods~\cite{kim2024promptkd,zhong2023e2s2,liu2022p,DBLP:conf/acl/LiL20/prefix}.
Accordingly, Table \ref{tab:module_abs} shows outcomes for different online modules applied during the distillation from GPT-2 XLarge to GPT-2 Large. 
The results confirm that enhancements in Forward KD are possible even with variations in component substitutions. This highlights that the strength of our method primarily derives from the online training approach rather than a specific component module. 
Ultimately, we selected LoRA as the online module for its superior performance.

\subsection{Further Analysis}\label{metric}

\begin{figure}
\centering
\begin{minipage}[t]{0.45\linewidth}
\centering\includegraphics[width=1\linewidth]{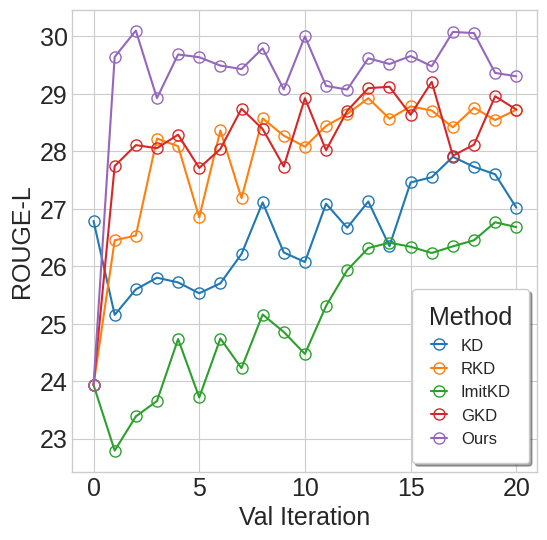}
\vspace{-1.5em}
    \caption{ROUGE-L scores for the validation set across the different methods.}
    \label{rouge_appendix}
\end{minipage}
\hspace{0.7em}
\begin{minipage}[t]{0.45\linewidth}
    \centering
\includegraphics[width=1\linewidth]{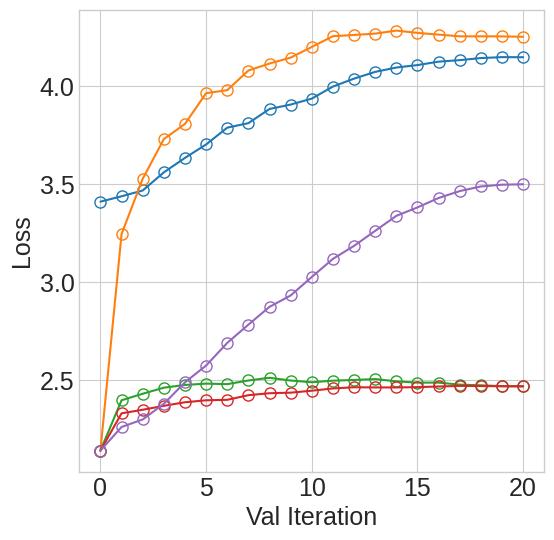}
\vspace{-1.5em}
    \caption{Plot of validation loss values across each validation iteration. 
    }
    \label{val_loss_appendix}
\end{minipage}
\end{figure}
\begin{figure}
\centering
\begin{minipage}[h]{0.45\linewidth}
\centering\includegraphics[width=1\linewidth]{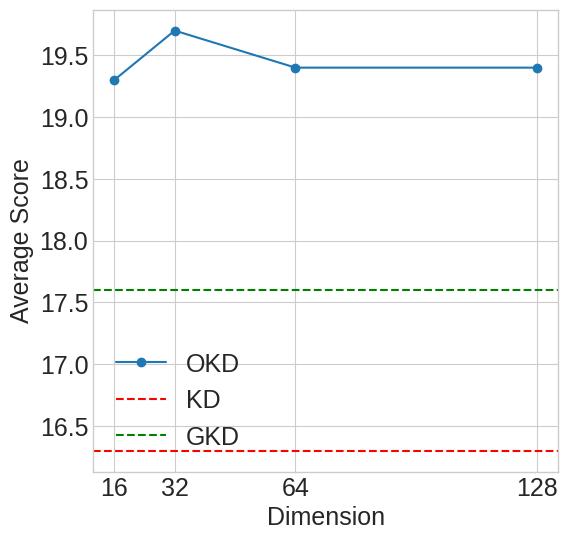}
\vspace{-1.5em}
    \caption{Effect of the dimension of the LoRA rank.}
    \label{dim_ablation}
\end{minipage}
\hspace{0.7em}
\begin{minipage}[h]{0.45\linewidth}
    \centering
\includegraphics[width=1\linewidth]{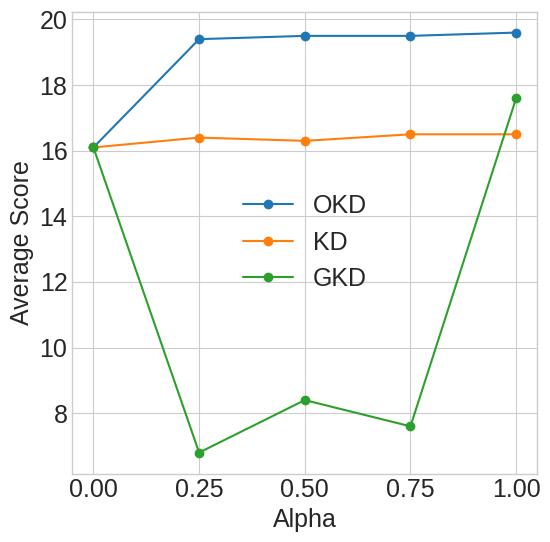}
    \vspace{-1.5em}
    \caption{Effect of the loss weight of the student.}
    \label{loss_weight}
\end{minipage}
\end{figure}

\begin{figure}[t]
    \centering
\includegraphics[width=0.8\linewidth]{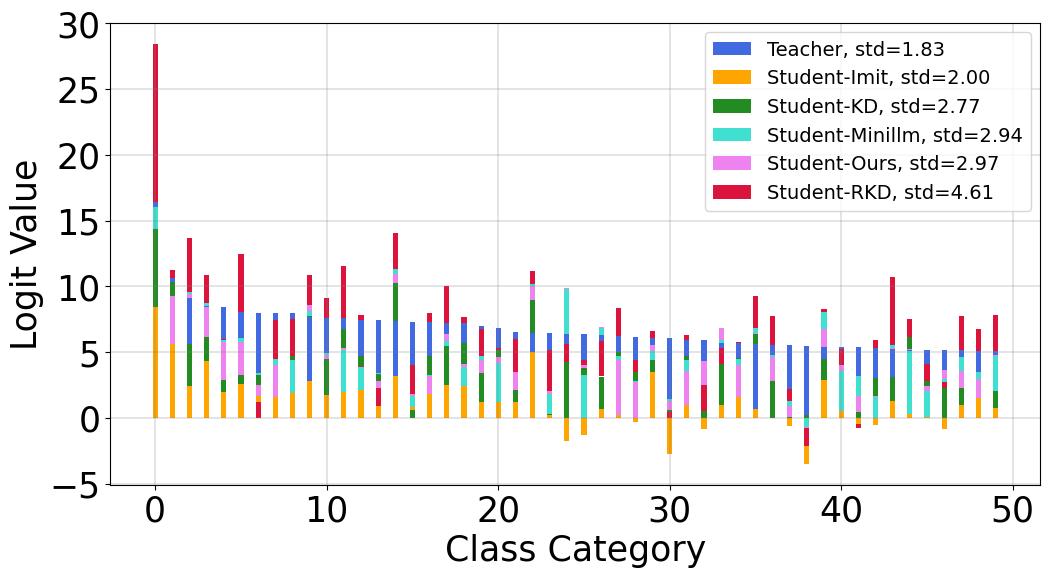}
    \caption{Example of ``\textbackslash n\textbackslash n Instruction:\textbackslash n What is the Masters?\textbackslash n\textbackslash n Response:\textbackslash n The Masters ''. The next word is ``Tournament''.}\label{example1}
\end{figure}
\begin{figure}[t]
    \centering
\includegraphics[width=0.8\linewidth]{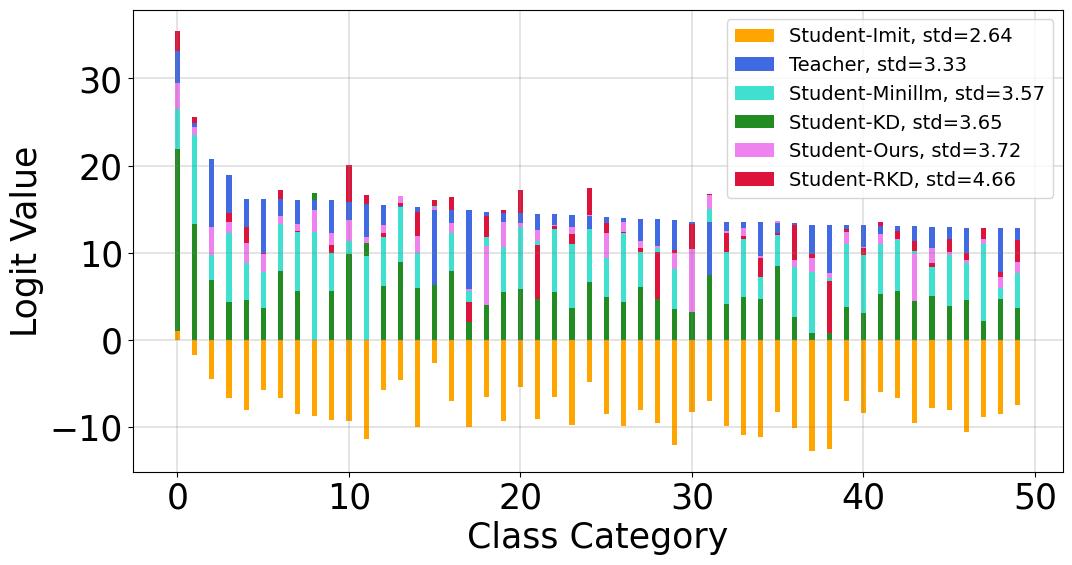}
    \caption{Example of ``\textbackslash n\textbackslash n Instruction: Which is a species of fish? Tope or Rope \textbackslash n\textbackslash n 
 Response: ''. The next word is ``Tope''.}\label{example2}
\end{figure}
\begin{figure}[t!]
    \centering
\includegraphics[width=0.8\linewidth]{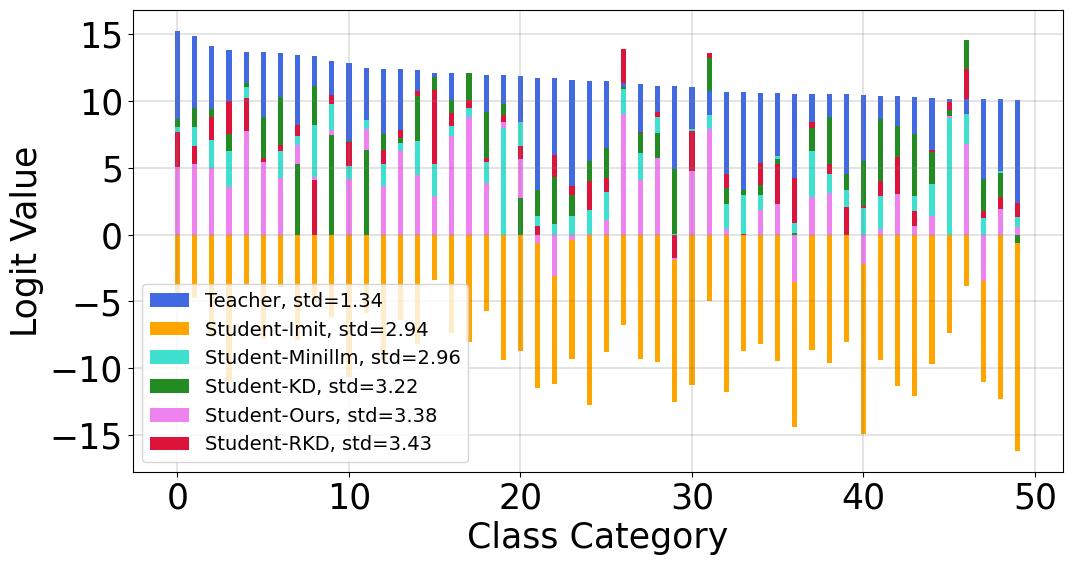}
    \caption{Example of ``\textbackslash n\textbackslash n Instruction:\textbackslash n What is a polygon?\textbackslash n\textbackslash n Response:\textbackslash n A polygon is a form in ``. The next word is ``Geometry''.}\label{example4}
\end{figure}

\paragraph{Validation RougeL and Loss}
We compare the validation set results for the GPT-2 XLarge to GPT-2 Large scale in Figures \ref{rouge_appendix} and \ref{val_loss_appendix}.
Our method achieves the highest RogueL scores and a rapid rise in validation set scores early in training, demonstrating rapid convergence and robust generalization capabilities.
Notably, our method significantly lowers the validation set loss (before 3 iterations), effectively mitigating this training-inference mismatch and achieving decent responses.
Additionally, our proposed method achieves the lowest error rate and the highest generation score, suggesting significant enhancements in distillation learning. 

\paragraph{LoRA Dimension}
As depicted in Figure \ref{dim_ablation}, we assess the impact of various scaling dimensions of the LoRA on NLG tasks, transitioning from GPT-2 XLarge to GPT-2 Base. 
The LoRA structure, as described in the methods section, consists of $w_{down}$, a non-linear transformation, and $w_{up}$. 
We modified the output dimension of $w_{down}$ and the input dimension of $w_{up}$. The dimensions tested—16, 64, and 128—do not appear to influence performance significantly. 
Following the guidelines in \cite{toward_unified_view_adapter}, we ultimately selected a dimension of 32, which is lower than the input dimension.

\paragraph{Loss Weight}
We further investigate whether students need to learn the true labels or if the teacher's distribution is sufficient. We utilize GPT-2 XLarge as the teacher and GPT-2 Base as the student, where $\alpha=0$ represents SFT, our findings are presented in Figure \ref{loss_weight}. The results indicate that optimal outcomes are achieved solely by learning the teacher's distribution, without any significant bias across the methods. Consequently, we adopted this setup for our experiments.
Additionally, our method demonstrated the highest consistency, highlighting its superior performance.

\subsection{Token Agreement Cases}
In summary, our method more faithfully mirrors the teacher's output in complex sentences and with rare words, suggesting a strong potential for distillation. We show examples of hard tokens in Figures ~\ref{example1}, \ref{example2}, and \ref{example4}.
We show the top 50 most probable predicted values for teachers.  We use  GPT-2 XLarge as a teacher and GPT-2 Base as a student. 
As we can see in the figures, it is evident that our approach closely mirrors the teacher's output in predicting rare words, while also avoiding overconfidence—specifically, it does not excessively exceed the teacher in TOP 1 predictions. Overall, our method demonstrates enhanced performance in predicting challenging words.


\section{Conclusion}
We systematically reviewed the characteristics and shortcomings of existing KD methods in LLM and found that fixed teacher distributions are a key issue preventing further improvements in current autoregressive language models.
Our approach minimizes distribution mismatch by integrating an online module, concurrently updating both the teacher's online module and the student model. Extensive testing across various generation tasks has proven OKD's superiority, which significantly enhances training efficiency and overall performance.

\bibliographystyle{IEEEtran}
\bibliography{custom.bib}

\begin{thebibliography}{10}
\providecommand{\url}[1]{#1}
\csname url@samestyle\endcsname
\providecommand{\newblock}{\relax}
\providecommand{\bibinfo}[2]{#2}
\providecommand{\BIBentrySTDinterwordspacing}{\spaceskip=0pt\relax}
\providecommand{\BIBentryALTinterwordstretchfactor}{4}
\providecommand{\BIBentryALTinterwordspacing}{\spaceskip=\fontdimen2\font plus
\BIBentryALTinterwordstretchfactor\fontdimen3\font minus \fontdimen4\font\relax}
\providecommand{\BIBforeignlanguage}[2]{{%
\expandafter\ifx\csname l@#1\endcsname\relax
\typeout{** WARNING: IEEEtran.bst: No hyphenation pattern has been}%
\typeout{** loaded for the language `#1'. Using the pattern for}%
\typeout{** the default language instead.}%
\else
\language=\csname l@#1\endcsname
\fi
#2}}
\providecommand{\BIBdecl}{\relax}
\BIBdecl

\bibitem{gpt3}
\BIBentryALTinterwordspacing
T.~B. Brown, B.~Mann, N.~Ryder, M.~Subbiah, J.~Kaplan, P.~Dhariwal, A.~Neelakantan, P.~Shyam, G.~Sastry, A.~Askell, S.~Agarwal, A.~Herbert{-}Voss, G.~Krueger, T.~Henighan, R.~Child, A.~Ramesh, D.~M. Ziegler, J.~Wu, C.~Winter, C.~Hesse, M.~Chen, E.~Sigler, M.~Litwin, S.~Gray, B.~Chess, J.~Clark, C.~Berner, S.~McCandlish, A.~Radford, I.~Sutskever, and D.~Amodei, ``Language models are few-shot learners,'' in \emph{NeurIPS}, 2020. [Online]. Available: \url{https://arxiv.org/abs/2005.14165}
\BIBentrySTDinterwordspacing

\bibitem{gpt4}
\BIBentryALTinterwordspacing
OpenAI, ``{GPT-4} technical report,'' in \emph{ArXiv}, 2023. [Online]. Available: \url{https://arxiv.org/abs/2303.08774}
\BIBentrySTDinterwordspacing

\bibitem{llama}
\BIBentryALTinterwordspacing
H.~Touvron, T.~Lavril, G.~Izacard, X.~Martinet, M.~Lachaux, T.~Lacroix, B.~Rozi{\`{e}}re, N.~Goyal, E.~Hambro, F.~Azhar, A.~Rodriguez, A.~Joulin, E.~Grave, and G.~Lample, ``Llama: Open and efficient foundation language models,'' in \emph{ArXiv}, 2023. [Online]. Available: \url{https://arxiv.org/abs/2302.13971}
\BIBentrySTDinterwordspacing

\bibitem{DBLP:journals/corr/HintonVD15}
\BIBentryALTinterwordspacing
G.~E. Hinton, O.~Vinyals, and J.~Dean, ``Distilling the knowledge in a neural network,'' in \emph{NeurIPS Workshop}, 2015. [Online]. Available: \url{https://arxiv.org/abs/1503.02531}
\BIBentrySTDinterwordspacing

\bibitem{10476767}
\BIBentryALTinterwordspacing
L.~Wu, H.~Lin, Z.~Gao, G.~Zhao, and S.~Z. Li, ``A teacher-free graph knowledge distillation framework with dual self-distillation,'' \emph{IEEE Transactions on Knowledge and Data Engineering}, vol.~36, no.~9, pp. 4375--4385, 2024. [Online]. Available: \url{https://ieeexplore.ieee.org/stamp/stamp.jsp?tp=&arnumber=10476767}
\BIBentrySTDinterwordspacing

\bibitem{10145833}
\BIBentryALTinterwordspacing
X.~Ouyang, Y.~Yang, W.~Zhou, Y.~Zhang, H.~Wang, and W.~Huang, ``Citytrans: Domain-adversarial training with knowledge transfer for spatio-temporal prediction across cities,'' \emph{IEEE Transactions on Knowledge and Data Engineering}, vol.~36, no.~1, pp. 62--76, 2024. [Online]. Available: \url{https://ieeexplore.ieee.org/document/10145833}
\BIBentrySTDinterwordspacing

\bibitem{PKD}
\BIBentryALTinterwordspacing
S.~Sun, Y.~Cheng, Z.~Gan, and J.~Liu, ``Patient knowledge distillation for {BERT} model compression,'' in \emph{EMNLP}, 2019. [Online]. Available: \url{https://arxiv.org/abs/1908.09355}
\BIBentrySTDinterwordspacing

\bibitem{DBLP:conf/cikm/RaoQQW0021}
\BIBentryALTinterwordspacing
J.~Rao, T.~Qian, S.~Qi, Y.~Wu, Q.~Liao, and X.~Wang, ``Student can also be a good teacher: Extracting knowledge from vision-and-language model for cross-modal retrieval,'' in \emph{CIKM}, 2021. [Online]. Available: \url{https://dl.acm.org/doi/abs/10.1145/3459637.3482194}
\BIBentrySTDinterwordspacing

\bibitem{zhong2024panda}
\BIBentryALTinterwordspacing
Q.~Zhong, L.~Ding, J.~Liu, B.~Du, and D.~Tao, ``Panda: Prompt transfer meets knowledge distillation for efficient model adaptation,'' \emph{IEEE Transactions on Knowledge and Data Engineering}, 2024. [Online]. Available: \url{https://ieeexplore.ieee.org/document/10475529}
\BIBentrySTDinterwordspacing

\bibitem{ProKD}
J.~Yim, D.~Joo, J.~Bae, and J.~Kim, ``A gift from knowledge distillation: Fast optimization, network minimization and transfer learning,'' in \emph{CVPR}, 2017.

\bibitem{KDCL}
\BIBentryALTinterwordspacing
Q.~Guo, X.~Wang, Y.~Wu, Z.~Yu, D.~Liang, X.~Hu, and P.~Luo, ``Online knowledge distillation via collaborative learning,'' in \emph{CVPR}, 2020. [Online]. Available: \url{https://openaccess.thecvf.com/content_CVPR_2020/papers/Guo_Online_Knowledge_Distillation_via_Collaborative_Learning_CVPR_2020_paper.pdf}
\BIBentrySTDinterwordspacing

\bibitem{DBLP:journals/corr/BahdanauCB14/self-attention}
\BIBentryALTinterwordspacing
D.~Bahdanau, K.~Cho, and Y.~Bengio, ``Neural machine translation by jointly learning to align and translate,'' in \emph{ICLR}, 2015. [Online]. Available: \url{https://arxiv.org/abs/1409.0473}
\BIBentrySTDinterwordspacing

\bibitem{DBLP:conf/nips/VaswaniSPUJGKP17}
\BIBentryALTinterwordspacing
A.~Vaswani, N.~Shazeer, N.~Parmar, J.~Uszkoreit, L.~Jones, A.~N. Gomez, L.~Kaiser, and I.~Polosukhin, ``Attention is all you need,'' in \emph{NeurIPS}, 2017. [Online]. Available: \url{https://arxiv.org/abs/1706.03762}
\BIBentrySTDinterwordspacing

\bibitem{bengio2015scheduled}
\BIBentryALTinterwordspacing
S.~Bengio, O.~Vinyals, N.~Jaitly, and N.~Shazeer, ``Scheduled sampling for sequence prediction with recurrent neural networks,'' \emph{NeurIPS}, vol.~28, 2015. [Online]. Available: \url{https://arxiv.org/abs/1506.03099}
\BIBentrySTDinterwordspacing

\bibitem{arora-etal-2022-exposure}
\BIBentryALTinterwordspacing
K.~Arora, L.~El~Asri, H.~Bahuleyan, and J.~Cheung, ``Why exposure bias matters: An imitation learning perspective of error accumulation in language generation,'' in \emph{Findings of the ACL}, S.~Muresan, P.~Nakov, and A.~Villavicencio, Eds.\hskip 1em plus 0.5em minus 0.4em\relax Dublin, Ireland: Association for Computational Linguistics, May 2022, pp. 700--710. [Online]. Available: \url{https://aclanthology.org/2022.findings-acl.58}
\BIBentrySTDinterwordspacing

\bibitem{agarwal2023gkd}
\BIBentryALTinterwordspacing
R.~Agarwal, N.~Vieillard, P.~Stanczyk, S.~Ramos, M.~Geist, and O.~Bachem, ``Gkd: Generalized knowledge distillation for auto-regressive sequence models,'' \emph{ICLR}, 2024. [Online]. Available: \url{https://arxiv.org/pdf/2306.13649.pdf}
\BIBentrySTDinterwordspacing

\bibitem{gu2023minillm}
\BIBentryALTinterwordspacing
Y.~Gu, L.~Dong, F.~Wei, and M.~Huang, ``Minillm: Knowledge distillation of large language models,'' in \emph{ICLR}, 2024. [Online]. Available: \url{https://arxiv.org/abs/2306.08543}
\BIBentrySTDinterwordspacing

\bibitem{wen-etal-2023-f}
\BIBentryALTinterwordspacing
Y.~Wen, Z.~Li, W.~Du, and L.~Mou, ``f-divergence minimization for sequence-level knowledge distillation,'' in \emph{ACL}, A.~Rogers, J.~Boyd-Graber, and N.~Okazaki, Eds.\hskip 1em plus 0.5em minus 0.4em\relax Toronto, Canada: Association for Computational Linguistics, Jul. 2023, pp. 10\,817--10\,834. [Online]. Available: \url{https://aclanthology.org/2023.acl-long.605}
\BIBentrySTDinterwordspacing

\bibitem{park2021learning}
\BIBentryALTinterwordspacing
D.~Y. Park, M.-H. Cha, D.~Kim, B.~Han \emph{et~al.}, ``Learning student-friendly teacher networks for knowledge distillation,'' \emph{NeurIPS}, 2021. [Online]. Available: \url{https://arxiv.org/abs/2102.07650}
\BIBentrySTDinterwordspacing

\bibitem{DBLP:conf/iccv/ZhuW21a}
\BIBentryALTinterwordspacing
Y.~Zhu and Y.~Wang, ``Student customized knowledge distillation: Bridging the gap between student and teacher,'' in \emph{ICCV}, 2021. [Online]. Available: \url{https://openaccess.thecvf.com/content/ICCV2021/papers/Zhu_Student_Customized_Knowledge_Distillation_Bridging_the_Gap_Between_Student_and_ICCV_2021_paper.pdf}
\BIBentrySTDinterwordspacing

\bibitem{RCO}
\BIBentryALTinterwordspacing
X.~Jin, B.~Peng, Y.~Wu, Y.~Liu, J.~Liu, D.~Liang, J.~Yan, and X.~Hu, ``Knowledge distillation via route constrained optimization,'' in \emph{ICCV}, 2019. [Online]. Available: \url{https://arxiv.org/pdf/1904.09149}
\BIBentrySTDinterwordspacing

\bibitem{DBLP:conf/aaai/MirzadehFLLMG20}
\BIBentryALTinterwordspacing
S.~Mirzadeh, M.~Farajtabar, A.~Li, N.~Levine, A.~Matsukawa, and H.~Ghasemzadeh, ``Improved knowledge distillation via teacher assistant,'' in \emph{AAAI}, 2020. [Online]. Available: \url{https://arxiv.org/abs/1902.03393}
\BIBentrySTDinterwordspacing

\bibitem{DBLP:journals/jmlr/RossB10}
S.~Ross and D.~Bagnell, ``Efficient reductions for imitation learning,'' in \emph{{AISTATS}}, ser. {JMLR} Proceedings, vol.~9.\hskip 1em plus 0.5em minus 0.4em\relax JMLR.org, 2010, pp. 661--668.

\bibitem{DBLP:conf/emnlp/LinWCL20}
\BIBentryALTinterwordspacing
A.~Lin, J.~Wohlwend, H.~Chen, and T.~Lei, ``Autoregressive knowledge distillation through imitation learning,'' in \emph{{EMNLP}}.\hskip 1em plus 0.5em minus 0.4em\relax Association for Computational Linguistics, 2020, pp. 6121--6133. [Online]. Available: \url{https://arxiv.org/abs/2009.07253}
\BIBentrySTDinterwordspacing

\bibitem{alpaca}
R.~Taori, I.~Gulrajani, T.~Zhang, Y.~Dubois, X.~Li, C.~Guestrin, P.~Liang, and T.~B. Hashimoto, ``Stanford alpaca: An instruction-following llama model,'' \url{https://github.com/tatsu-lab/stanford_alpaca}, 2023.

\bibitem{alpaca-gpt4}
\BIBentryALTinterwordspacing
B.~Peng, C.~Li, P.~He, M.~Galley, and J.~Gao, ``Instruction tuning with {GPT-4},'' in \emph{ArXiv}, 2023. [Online]. Available: \url{https://arxiv.org/abs/2304.03277}
\BIBentrySTDinterwordspacing

\bibitem{xu2023baize}
\BIBentryALTinterwordspacing
C.~Xu, D.~Guo, N.~Duan, and J.~McAuley, ``Baize: An open-source chat model with parameter-efficient tuning on self-chat data,'' in \emph{ArXiv}, 2023. [Online]. Available: \url{https://arxiv.org/abs/2304.01196}
\BIBentrySTDinterwordspacing

\bibitem{phoenix-2023}
\BIBentryALTinterwordspacing
Z.~Chen, F.~Jiang, J.~Chen, T.~Wang, F.~Yu, G.~Chen, H.~Zhang, J.~Liang, C.~Zhang, Z.~Zhang, J.~Li, X.~Wan, B.~Wang, and H.~Li, ``Phoenix: Democratizing chatgpt across languages,'' in \emph{ArXiv}, 2023. [Online]. Available: \url{https://arxiv.org/abs/2304.10453}
\BIBentrySTDinterwordspacing

\bibitem{wang2023far}
\BIBentryALTinterwordspacing
Y.~Wang, H.~Ivison, P.~Dasigi, J.~Hessel, T.~Khot, K.~R. Chandu, D.~Wadden, K.~MacMillan, N.~A. Smith, I.~Beltagy, and H.~Hajishirzi, ``How far can camels go? exploring the state of instruction tuning on open resources,'' in \emph{NeurIPS}, 2023. [Online]. Available: \url{https://arxiv.org/abs/2306.04751}
\BIBentrySTDinterwordspacing

\bibitem{flan2021}
\BIBentryALTinterwordspacing
J.~Wei, M.~Bosma, V.~Y. Zhao, K.~Guu, A.~W. Yu, B.~Lester, N.~Du, A.~M. Dai, and Q.~V. Le, ``Finetuned language models are zero-shot learners,'' in \emph{{ICLR}}, 2022. [Online]. Available: \url{https://openreview.net/forum?id=gEZrGCozdqR}
\BIBentrySTDinterwordspacing

\bibitem{supernli}
\BIBentryALTinterwordspacing
Y.~Wang, S.~Mishra, P.~Alipoormolabashi, Y.~Kordi, A.~Mirzaei, A.~Naik, A.~Ashok, A.~S. Dhanasekaran, A.~Arunkumar, D.~Stap, E.~Pathak, G.~Karamanolakis, H.~G. Lai, I.~Purohit, I.~Mondal, J.~Anderson, K.~Kuznia, K.~Doshi, K.~K. Pal, M.~Patel, M.~Moradshahi, M.~Parmar, M.~Purohit, N.~Varshney, P.~R. Kaza, P.~Verma, R.~S. Puri, R.~Karia, S.~Doshi, S.~K. Sampat, S.~Mishra, S.~R. A, S.~Patro, T.~Dixit, and X.~Shen, ``Super-naturalinstructions: Generalization via declarative instructions on 1600+ {NLP} tasks,'' in \emph{{EMNLP}}, 2022. [Online]. Available: \url{https://aclanthology.org/2022.emnlp-main.340.pdf}
\BIBentrySTDinterwordspacing

\bibitem{cot}
\BIBentryALTinterwordspacing
J.~Wei, X.~Wang, D.~Schuurmans, M.~Bosma, B.~Ichter, F.~Xia, E.~H. Chi, Q.~V. Le, and D.~Zhou, ``Chain-of-thought prompting elicits reasoning in large language models,'' in \emph{NeurIPS}, 2022. [Online]. Available: \url{https://openreview.net/pdf?id=_VjQlMeSB_J}
\BIBentrySTDinterwordspacing

\bibitem{flan_v2}
\BIBentryALTinterwordspacing
H.~W. Chung, L.~Hou, S.~Longpre, B.~Zoph, Y.~Tay, W.~Fedus, E.~Li, X.~Wang, M.~Dehghani, S.~Brahma, A.~Webson, S.~S. Gu, Z.~Dai, M.~Suzgun, X.~Chen, A.~Chowdhery, S.~Narang, G.~Mishra, A.~Yu, V.~Y. Zhao, Y.~Huang, A.~M. Dai, H.~Yu, S.~Petrov, E.~H. Chi, J.~Dean, J.~Devlin, A.~Roberts, D.~Zhou, Q.~V. Le, and J.~Wei, ``Scaling instruction-finetuned language models,'' in \emph{ArXiv}, 2022. [Online]. Available: \url{https://arxiv.org/abs/2210.11416}
\BIBentrySTDinterwordspacing

\bibitem{flan-moe}
\BIBentryALTinterwordspacing
S.~Shen, L.~Hou, Y.~Zhou, N.~Du, S.~Longpre, J.~Wei, H.~W. Chung, B.~Zoph, W.~Fedus, X.~Chen \emph{et~al.}, ``Flan-moe: Scaling instruction-finetuned language models with sparse mixture of experts,'' in \emph{ArXiv}, 2023. [Online]. Available: \url{https://arxiv.org/abs/2305.14705}
\BIBentrySTDinterwordspacing

\bibitem{DBLP:conf/acl/SuSKWHOYSZ023}
\BIBentryALTinterwordspacing
H.~Su, W.~Shi, J.~Kasai, Y.~Wang, Y.~Hu, M.~Ostendorf, W.~Yih, N.~A. Smith, L.~Zettlemoyer, and T.~Yu, ``One embedder, any task: Instruction-finetuned text embeddings,'' in \emph{{ACL} (Findings)}, 2023. [Online]. Available: \url{https://aclanthology.org/2023.findings-acl.71}
\BIBentrySTDinterwordspacing

\bibitem{meta-kd}
\BIBentryALTinterwordspacing
W.~Zhou, C.~Xu, and J.~J. McAuley, ``Bert learns to teach: Knowledge distillation with meta learning,'' in \emph{ACL}, 2022. [Online]. Available: \url{https://arxiv.org/abs/2106.04570}
\BIBentrySTDinterwordspacing

\bibitem{ding2021understanding}
\BIBentryALTinterwordspacing
L.~Ding, L.~Wang, X.~Liu, D.~F. Wong, D.~Tao, and Z.~Tu, ``Understanding and improving lexical choice in non-autoregressive translation,'' in \emph{ICLR}, 2021. [Online]. Available: \url{https://arxiv.org/abs/2012.14583}
\BIBentrySTDinterwordspacing

\bibitem{jing23seq}
\BIBentryALTinterwordspacing
J.~Li, P.~Han, X.~Ren, J.~Hu, L.~Chen, and S.~Shang, ``Sequence labeling with meta-learning,'' \emph{IEEE Transactions on Knowledge and Data Engineering}, vol.~35, no.~3, pp. 3072--3086, 2023. [Online]. Available: \url{https://doi.org/10.1109/TKDE.2021.3118469}
\BIBentrySTDinterwordspacing

\bibitem{rao2023DCD}
\BIBentryALTinterwordspacing
J.~Rao, L.~Ding, S.~Qi, M.~Fang, Y.~Liu, L.~Shen, and D.~Tao, ``Dynamic contrastive distillation for image-text retrieval,'' \emph{IEEE Transactions on Multimedia}, pp. 1--13, 2023. [Online]. Available: \url{https://ieeexplore.ieee.org/abstract/document/10102558/}
\BIBentrySTDinterwordspacing

\bibitem{CRD}
\BIBentryALTinterwordspacing
Y.~Tian, D.~Krishnan, and P.~Isola, ``Contrastive representation distillation,'' in \emph{ICLR}, 2020. [Online]. Available: \url{https://arxiv.org/abs/1910.10699}
\BIBentrySTDinterwordspacing

\bibitem{DBLP:conf/cvpr/ParkKLC19/RKD}
\BIBentryALTinterwordspacing
W.~Park, D.~Kim, Y.~Lu, and M.~Cho, ``Relational knowledge distillation,'' in \emph{CVPR}, 2019. [Online]. Available: \url{https://arxiv.org/abs/1904.05068}
\BIBentrySTDinterwordspacing

\bibitem{DML}
\BIBentryALTinterwordspacing
Y.~Zhang, T.~Xiang, T.~M. Hospedales, and H.~Lu, ``Deep mutual learning,'' in \emph{CVPR}, 2018. [Online]. Available: \url{https://arxiv.org/abs/1706.00384}
\BIBentrySTDinterwordspacing

\bibitem{pesf-kd}
\BIBentryALTinterwordspacing
J.~Rao, X.~Meng, L.~Ding, S.~Qi, X.~Liu, M.~Zhang, and D.~Tao, ``Parameter-efficient and student-friendly knowledge distillation,'' \emph{{IEEE} Trans. Multim.}, pp. 1--12, 2023. [Online]. Available: \url{https://arxiv.org/pdf/2205.15308}
\BIBentrySTDinterwordspacing

\bibitem{jing22nersurvey}
\BIBentryALTinterwordspacing
J.~Li, A.~Sun, J.~Han, and C.~Li, ``A survey on deep learning for named entity recognition,'' \emph{IEEE Transactions on Knowledge and Data Engineering (TKDE)}, vol.~34, no.~1, pp. 50--70, 2022. [Online]. Available: \url{https://doi.org/10.1109/TKDE.2020.2981314}
\BIBentrySTDinterwordspacing

\bibitem{ding2021rejuvenating}
\BIBentryALTinterwordspacing
L.~Ding, L.~Wang, X.~Liu, D.~F. Wong \emph{et~al.}, ``Rejuvenating low-frequency words: Making the most of parallel data in non-autoregressive translation,'' in \emph{ACL}.\hskip 1em plus 0.5em minus 0.4em\relax Association for Computational Linguistics, Aug. 2021, pp. 3431--3441. [Online]. Available: \url{https://aclanthology.org/2021.acl-long.266.pdf}
\BIBentrySTDinterwordspacing

\bibitem{ding2022redistributing}
\BIBentryALTinterwordspacing
L.~Ding, L.~Wang, S.~Shi, D.~Tao, and Z.~Tu, ``Redistributing low-frequency words: Making the most of monolingual data in non-autoregressive translation,'' in \emph{ACL}.\hskip 1em plus 0.5em minus 0.4em\relax Dublin, Ireland: Association for Computational Linguistics, May 2022, pp. 2417--2426. [Online]. Available: \url{https://aclanthology.org/2022.acl-long.172/}
\BIBentrySTDinterwordspacing

\bibitem{xiang2024dkdm}
\BIBentryALTinterwordspacing
Q.~Xiang, M.~Zhang, Y.~Shang, J.~Wu, Y.~Yan, and L.~Nie, ``Dkdm: Data-free knowledge distillation for diffusion models with any architecture,'' in \emph{Arxiv}, 2024. [Online]. Available: \url{https://arxiv.org/pdf/2409.03550}
\BIBentrySTDinterwordspacing

\bibitem{wu2024rethinking}
\BIBentryALTinterwordspacing
T.~Wu, C.~Tao, J.~Wang, Z.~Zhao, and N.~Wong, ``Rethinking kullback-leibler divergence in knowledge distillation for large language models,'' in \emph{ArXiv}, 2024. [Online]. Available: \url{https://arxiv.org/abs/2404.02657}
\BIBentrySTDinterwordspacing

\bibitem{kim2024promptkd}
\BIBentryALTinterwordspacing
G.~Kim, D.~Jang, and E.~Yang, ``Promptkd: Distilling student-friendly knowledge for generative language models via prompt tuning,'' in \emph{Arxiv}, 2024. [Online]. Available: \url{https://arxiv.org/abs/2402.12842}
\BIBentrySTDinterwordspacing

\bibitem{ko2024distillm}
\BIBentryALTinterwordspacing
J.~Ko, S.~Kim, T.~Chen, and S.-Y. Yun, ``Distillm: Towards streamlined distillation for large language models,'' in \emph{ICML}, 2024. [Online]. Available: \url{https://arxiv.org/abs/2402.03898}
\BIBentrySTDinterwordspacing

\bibitem{ji2023tailoring}
\BIBentryALTinterwordspacing
H.~Ji, P.~Ke, Z.~Hu, R.~Zhang, and M.~Huang, ``Tailoring language generation models under total variation distance,'' in \emph{ICLR}, 2023. [Online]. Available: \url{https://openreview.net/forum?id=VELL0PlWfc}
\BIBentrySTDinterwordspacing

\bibitem{rouge}
\BIBentryALTinterwordspacing
C.-Y. Lin, ``{ROUGE}: A package for automatic evaluation of summaries,'' in \emph{ACL}, 2004. [Online]. Available: \url{https://aclanthology.org/W04-1013}
\BIBentrySTDinterwordspacing

\bibitem{zhang-etal-2023-towards-understanding}
\BIBentryALTinterwordspacing
S.~Zhang, Y.~Liang, S.~Wang, Y.~Chen, W.~Han, J.~Liu, and J.~Xu, ``Towards understanding and improving knowledge distillation for neural machine translation,'' in \emph{ACL}, A.~Rogers, J.~Boyd-Graber, and N.~Okazaki, Eds.\hskip 1em plus 0.5em minus 0.4em\relax Toronto, Canada: Association for Computational Linguistics, Jul. 2023, pp. 8062--8079. [Online]. Available: \url{https://aclanthology.org/2023.acl-long.448}
\BIBentrySTDinterwordspacing

\bibitem{zhong2024revisiting}
\BIBentryALTinterwordspacing
Q.~Zhong, L.~Ding, L.~Shen, J.~Liu, B.~Du, and D.~Tao, ``Revisiting knowledge distillation for autoregressive language models,'' in \emph{ArXiv}, 2024. [Online]. Available: \url{https://arxiv.org/abs/2402.11890}
\BIBentrySTDinterwordspacing

\bibitem{radford2019language}
\BIBentryALTinterwordspacing
A.~Radford, J.~Wu, R.~Child, D.~Luan, D.~Amodei, I.~Sutskever \emph{et~al.}, ``Language models are unsupervised multitask learners,'' \emph{OpenAI blog}, vol.~1, no.~8, p.~9, 2019. [Online]. Available: \url{https://d4mucfpksywv.cloudfront.net/better-language-models/language_models_are_unsupervised_multitask_learners.pdf}
\BIBentrySTDinterwordspacing

\bibitem{zhang2024tinyllama}
\BIBentryALTinterwordspacing
P.~Zhang, G.~Zeng, T.~Wang, and W.~Lu, ``Tinyllama: An open-source small language model,'' in \emph{ArXiv}, 2024. [Online]. Available: \url{https://arxiv.org/abs/2401.02385}
\BIBentrySTDinterwordspacing

\bibitem{rao2022reproducibility}
\BIBentryALTinterwordspacing
J.~Rao, F.~Wang, L.~Ding, S.~Qi, Y.~Zhan, W.~Liu, and D.~Tao, ``Where does the performance improvement come from - a reproducibility concern about image-text retrieval,'' in \emph{SIGIR}, 2022. [Online]. Available: \url{https://arxiv.org/pdf/2203.03853}
\BIBentrySTDinterwordspacing

\bibitem{DatabricksBlog2023DollyV2}
\BIBentryALTinterwordspacing
M.~Conover, M.~Hayes, A.~Mathur, J.~Xie, J.~Wan, S.~Shah, A.~Ghodsi, P.~Wendell, M.~Zaharia, and R.~Xin, ``Free dolly: Introducing the world's first truly open instruction-tuned llm,'' 2023. [Online]. Available: \url{https://www.databricks.com/blog/2023/04/12/dolly-first-open-commercially-viable-\\instruction-tuned-llm}
\BIBentrySTDinterwordspacing

\bibitem{vicuna}
\BIBentryALTinterwordspacing
L.~Zheng, W.-L. Chiang, Y.~Sheng, S.~Zhuang, Z.~Wu, Y.~Zhuang, Z.~Lin, Z.~Li, D.~Li, E.~P. Xing, H.~Zhang, J.~E. Gonzalez, and I.~Stoica, ``Judging llm-as-a-judge with mt-bench and chatbot arena,'' in \emph{ArXiv}, 2023. [Online]. Available: \url{https://arxiv.org/abs/2306.05685}
\BIBentrySTDinterwordspacing

\bibitem{wang-etal-2023-self-instruct}
\BIBentryALTinterwordspacing
Y.~Wang, Y.~Kordi, S.~Mishra, A.~Liu, N.~A. Smith, D.~Khashabi, and H.~Hajishirzi, ``Self-instruct: Aligning language models with self-generated instructions,'' in \emph{ACL}.\hskip 1em plus 0.5em minus 0.4em\relax Toronto, Canada: Association for Computational Linguistics, Jul. 2023, pp. 13\,484--13\,508. [Online]. Available: \url{https://aclanthology.org/2023.acl-long.754}
\BIBentrySTDinterwordspacing

\bibitem{deepspeed}
\BIBentryALTinterwordspacing
J.~Rasley, S.~Rajbhandari, O.~Ruwase, and Y.~He, ``Deepspeed: System optimizations enable training deep learning models with over 100 billion parameters,'' in \emph{{KDD}}, 2020. [Online]. Available: \url{https://dl.acm.org/doi/10.1145/3394486.3406703}
\BIBentrySTDinterwordspacing

\bibitem{zero}
\BIBentryALTinterwordspacing
S.~Rajbhandari, J.~Rasley, O.~Ruwase, and Y.~He, ``Zero: memory optimizations toward training trillion parameter models,'' in \emph{{SC}}.\hskip 1em plus 0.5em minus 0.4em\relax {IEEE/ACM}, 2020, p.~20. [Online]. Available: \url{https://arxiv.org/abs/1910.02054}
\BIBentrySTDinterwordspacing

\bibitem{lora}
\BIBentryALTinterwordspacing
E.~J. Hu, Y.~Shen, P.~Wallis, Z.~Allen{-}Zhu, Y.~Li, S.~Wang, and W.~Chen, ``Lora: Low-rank adaptation of large language models,'' in \emph{ICLR}, 2022. [Online]. Available: \url{https://arxiv.org/abs/2106.09685}
\BIBentrySTDinterwordspacing

\bibitem{zhong2023e2s2}
\BIBentryALTinterwordspacing
Q.~Zhong, L.~Ding, J.~Liu, B.~Du, and D.~Tao, ``E2s2: Encoding-enhanced sequence-to-sequence pretraining for language understanding and generation,'' \emph{IEEE Transactions on Knowledge and Data Engineering}, 2023. [Online]. Available: \url{https://ieeexplore.ieee.org/document/10363656}
\BIBentrySTDinterwordspacing

\bibitem{liu2022p}
\BIBentryALTinterwordspacing
X.~Liu, K.~Ji, Y.~Fu, W.~Tam, Z.~Du, Z.~Yang, and J.~Tang, ``P-tuning: Prompt tuning can be comparable to fine-tuning across scales and tasks,'' in \emph{ACL}, 2022. [Online]. Available: \url{https://aclanthology.org/2022.acl-short.8}
\BIBentrySTDinterwordspacing

\bibitem{DBLP:conf/acl/LiL20/prefix}
\BIBentryALTinterwordspacing
X.~L. Li and P.~Liang, ``Prefix-tuning: Optimizing continuous prompts for generation,'' in \emph{ACL}, 2021. [Online]. Available: \url{https://aclanthology.org/2021.acl-long.353/}
\BIBentrySTDinterwordspacing

\bibitem{toward_unified_view_adapter}
\BIBentryALTinterwordspacing
J.~He, C.~Zhou, X.~Ma, T.~Berg{-}Kirkpatrick, and G.~Neubig, ``Towards a unified view of parameter-efficient transfer learning,'' in \emph{ICLR}, 2022. [Online]. Available: \url{https://arxiv.org/abs/2110.04366}
\BIBentrySTDinterwordspacing

\end{thebibliography}

\ifCLASSOPTIONcompsoc
  \section*{Acknowledgments}
\else
  \section*{Acknowledgment}
\fi

This work was supported in part by the National Natural Science Foundation of China (Grant No. 62206076), Guangdong Basic and Applied Basic Research Foundation (Grant No. 2024A1515011491), Shenzhen Science and Technology Program (Grant Nos. ZDSYS20230626091203008 and KJZD20231023094700001), and Shenzhen College Stability Support Plan (Grant Nos. GXWD20220811173340003 and GXWD20220817123150002).

\ifCLASSOPTIONcaptionsoff
  \newpage
\fi

\end{document}